\documentclass[letterpaper]{article} % DO NOT CHANGE THIS
\usepackage{aaai2026}  % DO NOT CHANGE THIS
\usepackage{times}  % DO NOT CHANGE THIS
\usepackage{helvet}  % DO NOT CHANGE THIS
\usepackage{courier}  % DO NOT CHANGE THIS
\usepackage[hyphens]{url}  % DO NOT CHANGE THIS
\usepackage{graphicx} % DO NOT CHANGE THIS
\usepackage{natbib}  % DO NOT CHANGE THIS AND DO NOT ADD ANY OPTIONS TO IT
\usepackage{caption} % DO NOT CHANGE THIS AND DO NOT ADD ANY OPTIONS TO IT
\usepackage{algorithm}
\usepackage{algorithmic}
\usepackage{xcolor}
\usepackage{amsmath}
\usepackage[most]{tcolorbox}

\newcommand{\method}{\textsc{Jupiter}}

\usepackage{newfloat}
\usepackage{listings}
\DeclareCaptionStyle{ruled}{labelfont=normalfont,labelsep=colon,strut=off} % DO NOT CHANGE THIS
\floatstyle{ruled}
\newfloat{listing}{tb}{lst}{}
\floatname{listing}{Listing}
\nocopyright

\usepackage{multirow}
\usepackage{array}
\usepackage{booktabs}

\title{\method: Enhancing LLM Data Analysis Capabilities via \\ Notebook and Inference-Time Value-Guided Search}
\author {
    Shuocheng Li\textsuperscript{\rm 1}\thanks{Work done during internship at Microsoft. This paper has been accepted to AAAI 2026.},
    Yihao Liu\textsuperscript{\rm 1}\footnotemark[1],
    Silin Du\textsuperscript{\rm 2}\footnotemark[1],
    Wenxuan Zeng\textsuperscript{\rm 1}\footnotemark[1],
    Zhe Xu\textsuperscript{\rm 1}\footnotemark[1],\\
    Mengyu Zhou\textsuperscript{\rm 3}\thanks{Corresponding author: \texttt{mengyu.chou@gmail.com}.},
    Yeye He\textsuperscript{\rm 3},
    Haoyu Dong\textsuperscript{\rm 3},
    Shi Han\textsuperscript{\rm 3},
    Dongmei Zhang\textsuperscript{\rm 3}
}

\affiliations{
    \textsuperscript{\rm 1}Peking University\quad\textsuperscript{\rm 2}Stanford University\quad\textsuperscript{\rm 3}Microsoft
}

\usepackage{bibentry}
\begin{document}

\maketitle

\maketitle

\begin{abstract}
Large language models (LLMs) have shown great promise in automating data science workflows. However, existing models still struggle with multi-step reasoning and tool use, limiting their effectiveness on complex data analysis tasks.
To address this limitation, we propose a scalable pipeline that extracts high-quality, tool-based data analysis tasks and their executable multi-step solutions from real-world Jupyter notebooks and associated data files.
Using this pipeline, we introduce NbQA, a large-scale dataset of standardized task–solution pairs that reflect authentic tool-use patterns in practical data science scenarios.
To further enhance the multi-step reasoning capabilities, we present \method, a framework that formulates data analysis as a search problem and applies Monte Carlo Tree Search (MCTS) to generate diverse solution trajectories for value model learning.
During inference, \method~combines the value model and
node visit counts to efficiently collect executable multi-step plans with minimal search steps.
Experimental results show that Qwen2.5-7B and 14B-Instruct models on NbQA solve 77.82\% and 86.38\% of tasks on InfiAgent-DABench, respectively—matching or surpassing GPT-4o and advanced agent frameworks. Further evaluations demonstrate improved generalization and stronger tool-use reasoning across diverse multi-step reasoning tasks. Code and data are available at \url{https://github.com/microsoft/Jupiter}.
\end{abstract}


\section{Introduction}

\begin{figure}[!tb]
    \centering
    \includegraphics[width=0.95\linewidth]{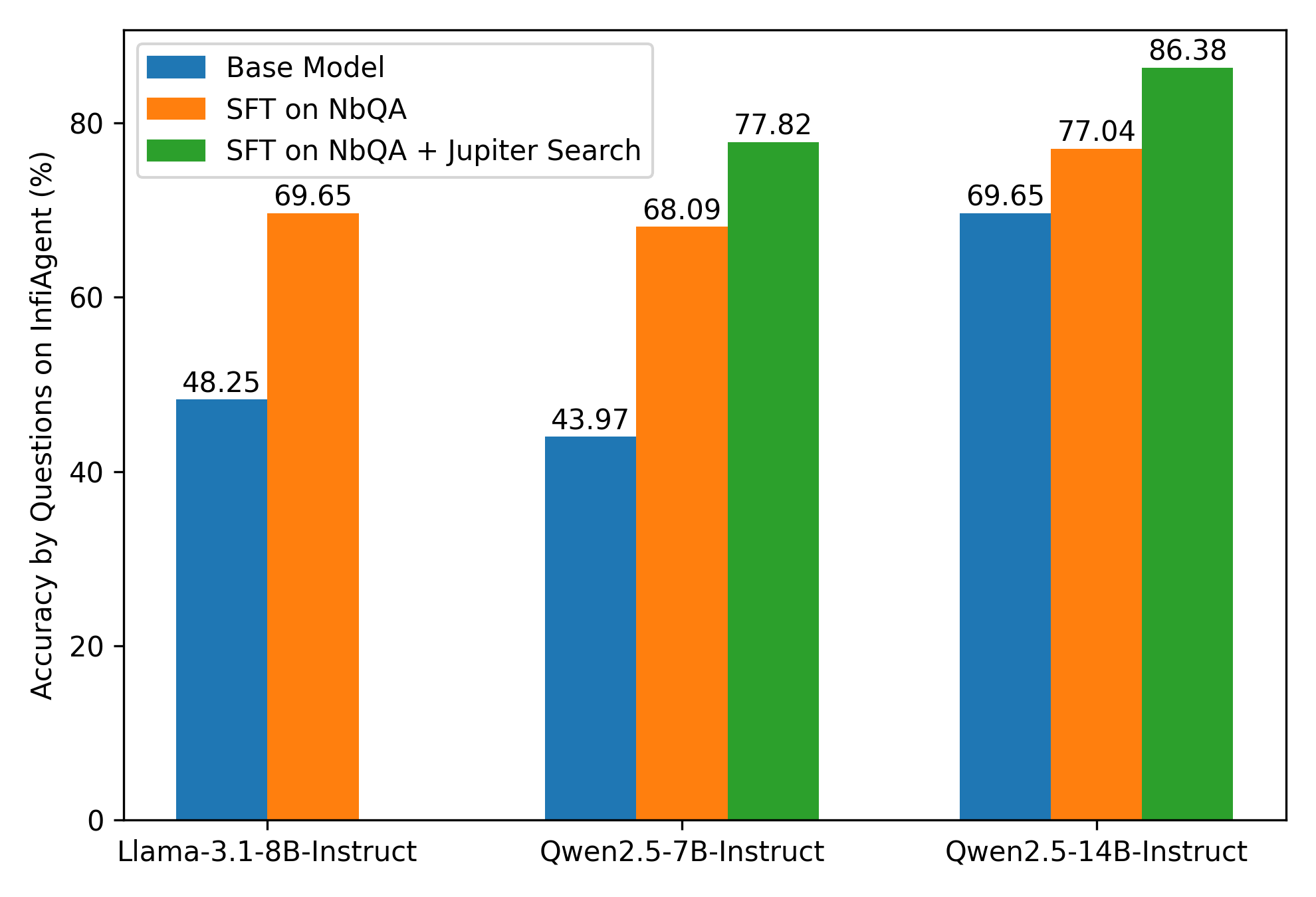}
    \caption{Accuracy by Questions (\%) across different models on InfiAgent-DABench under three settings: base model, SFT on \textbf{NbQA}, and SFT on \textbf{NbQA} + \textsc{Jupiter} search, demonstrating the base models suffer from accurate multi-step data analysis.}
    \label{fig:improvements}
\end{figure}

Data analysis is the process of extracting knowledge and insights from data \cite{Donoho201750YO}. A complete data analysis workflow often involves multi-step derivations and decisions where intermediate steps are highly interdependent. This process demands that practitioners continuously explore and iteratively optimize their approaches, guided by feedback from their decisions. Furthermore, addressing specific scenarios in data analysis frequently requires domain knowledge and specialized tools. Therefore, designing a correct and practical data analysis workflow is a challenging and complex task, posing significant obstacles to achieving fully automated solutions \cite{De_Bie_2022}.

Recent advancements in large language models (LLMs) and LLM-based agents \cite{cheng2023gpt4gooddataanalyst, qiao2024taskweavercodefirstagentframework, hollmann2023largelanguagemodelsautomated, zhang2024usinglargelanguagemodels} have demonstrated significant potential in automating certain data science tasks, such as feature engineering \cite{hollmann2023largelanguagemodelsautomated}, data visualization \cite{yang2024matplotagentmethodevaluationllmbased}, and model selection \cite{shen2023hugginggptsolvingaitasks}.However, these methods typically focus on only a single stage of the data analysis process, without considering the interdependencies across real-world data science tasks. Some agent-based systems (e.g., Data Interpreter \cite{hong2024datainterpreterllmagent}, AutoGen \cite{wu2023autogenenablingnextgenllm}, Taskweaver \cite{qiao2024taskweavercodefirstagentframework}) have attempted to provide more comprehensive solutions, but these approaches often rely on proprietary commercial models, complex task decomposition, and intricate system designs. Even the most advanced models—open or proprietary—still face significant limitations in data understanding and multi-step reasoning with tools \cite{hu2024infiagentdabenchevaluatingagentsdata}.

In this work, we aim to improve models' multi-step data analysis capabilities by combining multi-step solutions extracted from large-scale notebook corpora with inference-time value-guided search. Specifically, we first construct a task multi-step solution dataset named \textbf{NbQA} from a large collection of Jupyter notebooks, which inherently record multi-step problem-solving and reasoning processes as experts iteratively build and execute code.

Specifically, we first use the GitHub API \cite{githubrestapi} to crawl approximately 1.6M Jupyter notebooks and 3.2M associated data dependency files. After coarse filtering for valid structure, successful execution, sufficient data complexity and code, we further diversify the dataset by limiting the number of notebooks per repository and data source. All notebooks are then automatically evaluated for quality by GPT-4o mini, which also identifies and classifies the machine learning models present; only high-scoring samples involving common algorithms are retained. For fine-grained processing, GPT-4o is used to extract high-quality, evaluable task–solution pairs across diverse analysis types such as summary statistics, distribution analysis, and feature engineering, each annotated with explicit constraints and standardized output formats for automatic evaluation and value model training. This process yields \textbf{38,635} task–solution pairs in NbQA, of which \textbf{6,845} include complete, fully interactive data dependencies. Fine-tuning both 7B and 14B open-source models on 8,975 samples from NbQA demonstrates substantial improvements on InfiAgent-DABench, with 7B models achieving over \textbf{21\%} absolute accuracy improvement on InfiAgent-DABench \cite{hu2024infiagentdabenchevaluatingagentsdata}, demonstrating NbQA’s effectiveness for enhancing multi-step data analysis and reasoning.

To further enhance the model's ability to perform multi-step reasoning, we further propose the \method~framework, which formulates data analysis as a state-level search problem grounded in the Jupyter notebook paradigm. The search tree’s root node corresponds directly to the original data analysis problem to be solved. Each subsequent node in the tree represents a notebook state, including the accumulated model-generated thought–action pairs and their Jupyter kernel execution results, while leaf nodes denote either terminal error or candidate answer. At each step, \method~selects a node to expand by sampling the next thought–action based on the current notebook context. After a limited number of interactions, the final answer is selected from the candidate answer nodes collected during the search. 

Specifically, the subset of NbQA tasks with fully interactive data dependencies enables us to use a fine-tuned model to simulate and collect detailed multi-step reasoning trajectories via MCTS. For each task, we construct an MCTS search tree, where each trajectory terminates at either a correct answer or an error node, with rewards assigned and backpropagated through the tree. After sufficient simulations, the Q-value estimates for each node become more accurate. These collected trajectories—including both successful and failed solution paths—are then used to train value models, which guide inference-time search. During inference, we remove the exploration term in the PUCT algorithm \cite{Rosin2011MultiarmedBW}, relying primarily on node visit counts and value model estimates for tree expansion. On InfiAgent-DABench, our value-guided search enables the fine-tuned 7B and 14B models to solve 77.82\% and 86.38\% of tasks after 40 iterations, respectively—where the 14B model surpasses direct inference with models such as GPT-4o and GPT-4o mini, and advanced agent architectures using these same base models. 
We further validated the generalization and out-of-distribution capabilities of our trained value models on DSBench \cite{jing2025dsbenchfardatascience} and AIME \cite{opencompass2025aime}, demonstrating effectiveness across diverse data analysis tasks as well as improved numerical computation and multi-step tool-use reasoning in mathematical scenarios.

In summary, the contributions of our work are as follows:
\begin{itemize}
\item We propose an automated pipeline for extracting high-quality data analysis tasks and their multi-step solutions from large-scale Jupyter notebooks and associated data files. Using this pipeline, we construct \textbf{NbQA}, a large-scale, high-quality dataset of standardized task–solution pairs. NbQA covers diverse, practical data analysis scenarios, supporting both supervised fine-tuning and value model training.

\item We propose \method, a general framework that formulates notebook-based data analysis as a sequential decision-making search problem. \method~leverages a value model trained from interaction trajectories to guide inference-time search, efficiently discovering diverse candidate solutions under limited interaction steps.

\item We conduct comprehensive experiments across data analysis and out-of-domain benchmarks. Our approach significantly improves multi-step reasoning, tool-use, and generalization abilities of 7B and 14B models, matching or surpassing strong commercial models such as GPT-4o on InfiAgent-DABench, and demonstrating strong transferability on DSBench and AIME.
\end{itemize}

\section{Construction of the NbQA Dataset}

\begin{figure}[!tb]
    \centering
    \includegraphics[width=\linewidth]{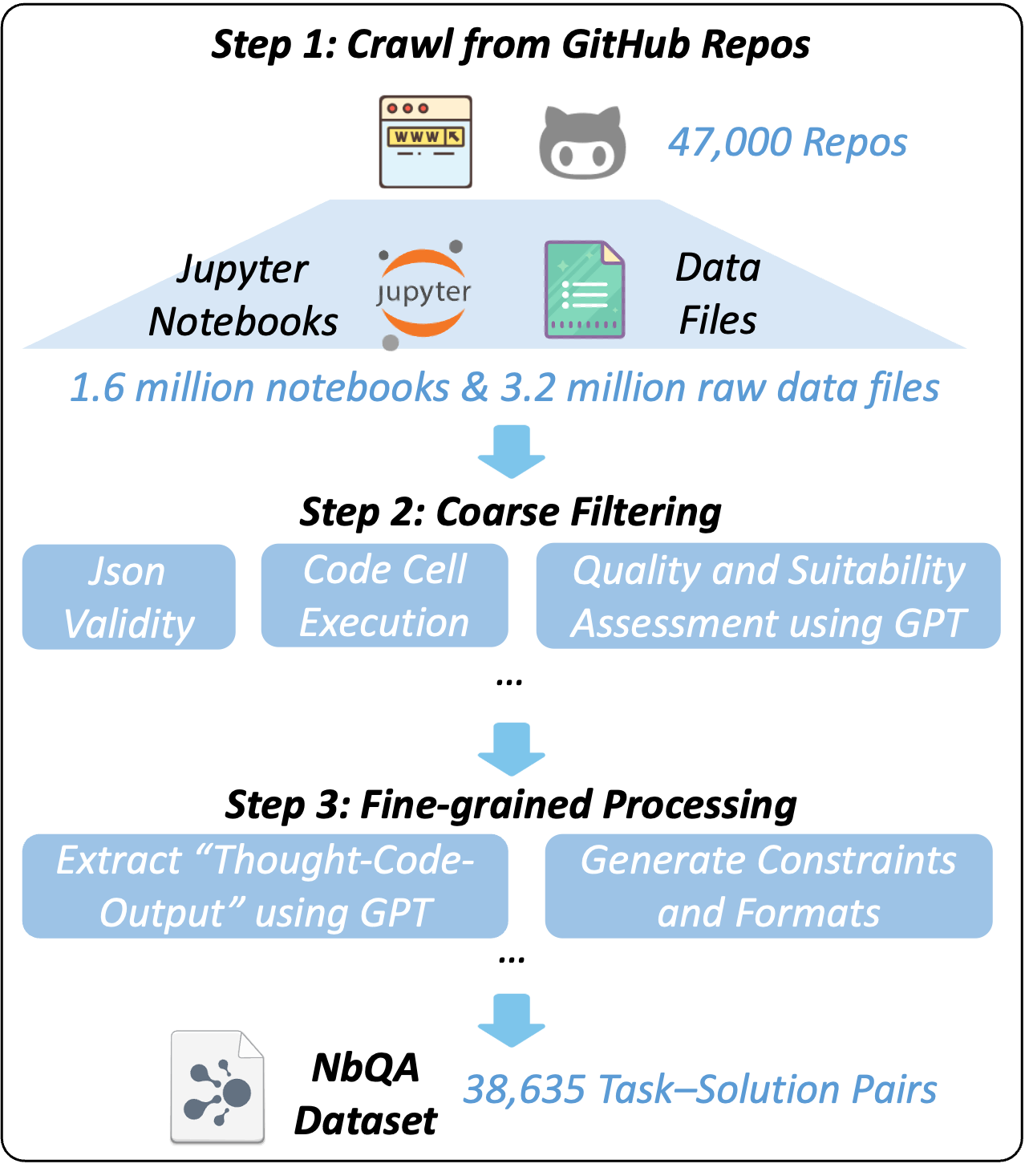}
    \caption{Construction of the NbQA dataset.}
    \label{fig:dataset}
\end{figure}

In this section, we describe the workflow used to construct the NbQA dataset (see Figure \ref{fig:dataset}). Details of the prompt design can be found in the supplementary materials.

\subsection{Crawl Notebooks on GitHub}

We first use the GitHub API to collect Jupyter Notebooks with the \texttt{.ipynb} suffix from open-source repositories, identify data loading functions via regular expressions, and extract the required data files. We then search for and download these files from each repository’s working directory. In total, we collect 1.6 million notebooks and 3.2 million raw data files from about 47{,}000 repositories. Notebooks with missing data dependencies are still retained for extracting task-solution pairs for supervised fine-tuning.

\subsection{Coarse Filtering}

To ensure data quality and reliability, we first validated all crawled Jupyter notebooks by checking their underlying JSON structure. We excluded notebooks that were not executed in code-cell order, contained unexecuted cells, or raised unhandled execution errors. To prevent data contamination and test set leakage, we performed keyword matching on filenames and notebook content to remove any notebooks referencing common teaching or competition datasets, such as Iris~\cite{iris_kaggle}, Titanic~\cite{titanic_kaggle}, Breast Cancer~\cite{breast_cancer_kaggle}, Boston Housing~\cite{boston_housing_kaggle}, and Wine Quality~\cite{wine_quality_kaggle}. We also removed notebooks involving overly simple tasks, such as those processing data files with fewer than 20 rows or containing fewer than 40 lines of code. 

As an additional quality control step, we prompted GPT-4o mini to assign a score from 1 to 5 for each notebook, based on factors such as code correctness and structure, data relevance and complexity, and the appropriate use of standard Python libraries. Only notebooks with a score of 3 or higher were retained.

To further ensure dataset quality and diversity, we applied additional filtering and deduplication strategies. We limited each GitHub repository to at most one retained notebook if its data files were incomplete, and at most two if complete. We also grouped notebooks by the hash values of their data dependency files and retained only those groups where all notebooks originated from the same repository, keeping at most two notebooks per group. Finally, to focus on classical data analysis tasks, we filtered out all notebooks involving neural networks, pre-trained models, or GPU-based computation.

\subsection{Fine-grained Processing}

To ensure that the dataset is thematically focused and structurally consistent, we begin by using GPT-4o mini to identify the machine learning models present in each notebook. For notebooks involving machine learning, we retain only those that utilize one or more of 21 commonly used classical models (e.g., Logistic Regression, Support Vector Machine, Linear Regression), which collectively cover major machine learning tasks.

We then define eight categories of data analysis tasks: Summary Statistics, Distribution Analysis, Correlation Analysis, Outlier Detection, Comprehensive Data Preprocessing, Feature Engineering, Machine Learning, and Visualization. Guided by examples, we instruct GPT-4o to extract 1 to 3 representative subtasks from each notebook, along with their corresponding answers and task types. To ensure answer verifiability, we require that, except for visualization tasks, all numeric answers must be directly extractable from the notebook's code outputs. The prompt also enforces strict constraints: extracted tasks must be well-defined, free of contextual assumptions, and must not merely restate code execution results.

To further address potential issues such as vague task descriptions or inconsistent answer formats, we instruct the model to supplement each task with explicit solution constraints and standardized output formats, and to present answers in the form of \texttt{@answer name[answer]} labels to facilitate automatic evaluation. We then prompt the model to generate multi-step solutions for each task. Unlike previous synthetic data approaches that rely on LLM-generated solutions, our method does not generate answers from scratch but instead extracts relevant code and outputs directly from the original notebook. The model is strictly prohibited from fabricating outputs or skipping steps, ensuring that each solution —comprising multiple \texttt{thought-code-output} blocks and a final \texttt{thought-label} block—faithfully reflects the actual analysis workflow.

Finally, we apply automated review using GPT-4o mini to filter out tasks with mismatched answers or vague constraints. After all filtering steps, the final NbQA dataset contains 38,635 task–solution pairs, of which 6,845 are associated with complete and fully interactive data dependency files and exhibit low randomness suitable for automatic evaluation.

\begin{figure}[!tb]
    \centering
    \includegraphics[width=\linewidth]{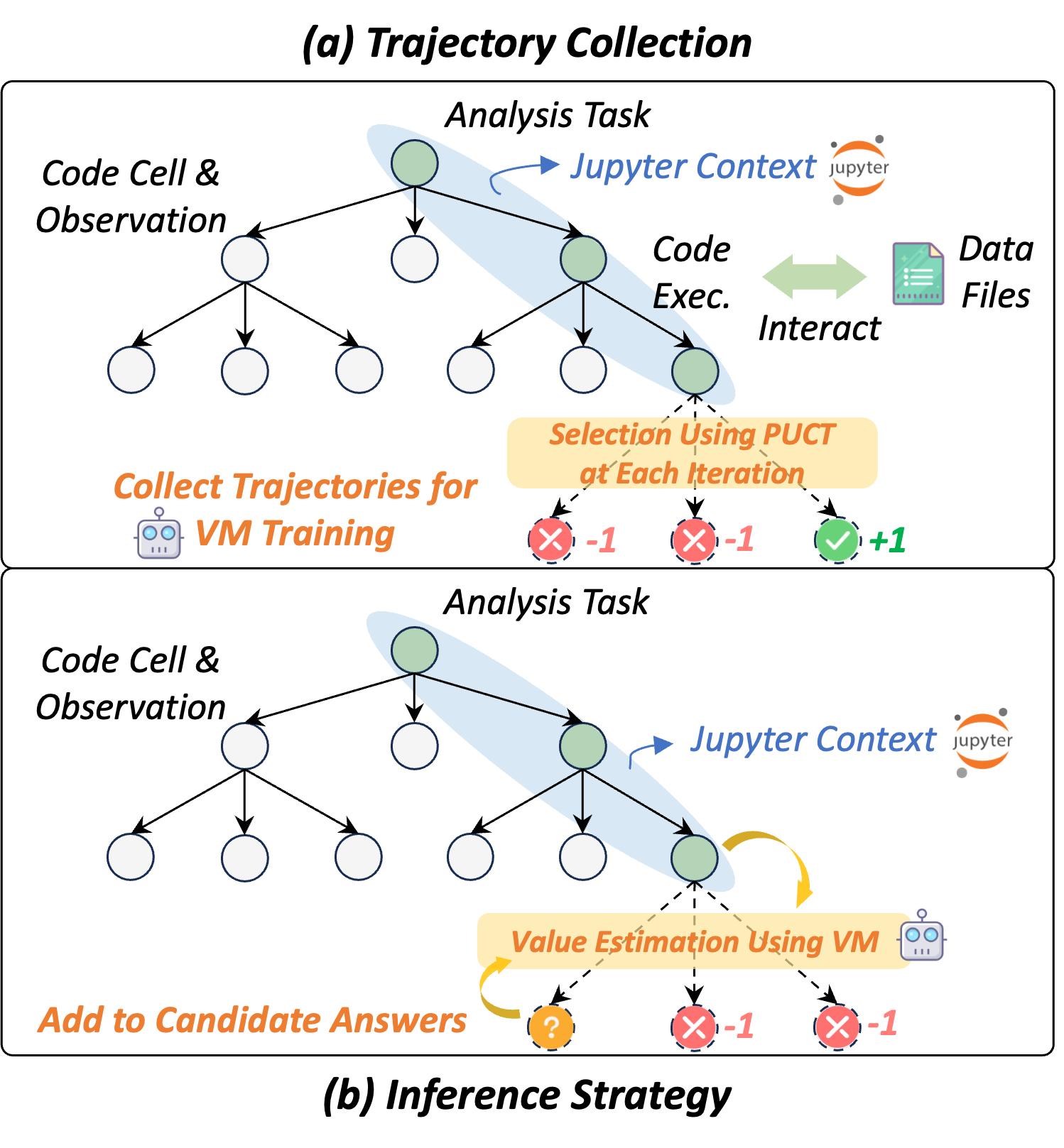}
    \caption{Overview of \method, including (a) trajectory collection and (b) inference strategy. The collected trajectories are used for training the value model (VM), and VM is used for value estimation during inference.}
    \label{fig:framework}
\end{figure}

\section{\method~Framework}

In this section, we demonstrate how the \method~framework transforms solving data analysis problems using notebooks into a search problem. The full procedure is illustrated in Figure~\ref{fig:framework} and detailed in Algorithm~\ref{alg:jupiter-mcts}. Then we briefly describe the structure and training process of the value model.

\subsection{Problem Formulation}

\begin{algorithm}[ht]
\caption{\method~for Notebook-based Data Analysis}
\label{alg:jupiter-mcts}
\begin{algorithmic}[1]
\REQUIRE Initial notebook state $s_0$, value model $V_\phi$ (optional), search budget $N$, number of expansions per step $K$
\STATE Initialize search tree with root node $s_0$
\FOR{iteration $=1$ \TO $N$}
    \IF{Trajectory Collection Phrase}
        \STATE Select node to expand using PUCT ($c_{puct}>0$)
    \ELSE
        \STATE Select node to expand using PUCT ($c_{puct}=0$)
    \ENDIF
    \STATE Sample $K$ candidate thought-action pairs from LLM
    \FOR{each candidate}
        \STATE Execute code in notebook context, obtain output
        \STATE Expand notebook content to create a new child node
        \IF{Trajectory Collection Phrase}
            \IF{Child node is terminal}
                \STATE Assign reward ($+1$ if correct answer, $-1$ if error/invalid)
            \ELSE
                \STATE Estimate value using $V_\phi$ if available
            \ENDIF
        \ELSE
            \STATE Estimate value using $V_\phi$ for this child node
        \ENDIF
        \STATE Backpropagate reward or value along the path
    \ENDFOR
\ENDFOR
\STATE \textbf{Return} candidate answer nodes collected during search
\end{algorithmic}
\end{algorithm}

We formalize multi-step data analysis as a sequential decision-making search problem. The process is modeled as a tree, where the root node represents the initial notebook state $s_0$—including the data analysis problem to be solved—and each edge corresponds to the generation and execution of a code cell. Each node in the tree records the current thought, code, and the resulting output, while the path from the root to any node forms the complete notebook context at that step.

At each search step, the agent selects a node to expand using the PUCT algorithm and samples $K$ candidate thought-action pairs from the language model. Each candidate is executed in the current notebook context to create a new child node. Terminal nodes represent answers or error states, while intermediate nodes can be further expanded. Rewards are backpropagated along the tree to affect the search process. Specifically, for each node $s'$, the PUCT score is calculated as follows:
\begin{equation}  
    \mathrm{PUCT}(s') = Q(s') + c_{\mathrm{puct}} \cdot P(s') \cdot \frac{\sqrt{N_{\text{parent}}(s')}}{1 + N(s')}
\end{equation}
where $Q(s')$ is the average value estimate of node $s'$, $P(s')$ is the prior probability assigned by the language model, $N_{\text{parent}}(s')$ is the visit count of the parent node, $N(s')$ is the visit count of node $s'$ itself, and $c_{\mathrm{puct}}$ is a hyperparameter controlling the tradeoff between exploitation term $Q(s')$ and exploration term $P(s') \cdot \frac{\sqrt{N_{\text{parent}}(s')}}{1 + N(s')}$. The node with the highest PUCT score is selected for expansion at each step.

During the trajectory collection phase, terminal nodes are assigned positive or negative rewards for correctness. For each node, its $Q(s')$ is estimated by averaging the returns from the observed rewards of its descendant nodes during the search. The collected trajectories, along with these Q values, are then used to train a value model that predicts the expected return for each notebook state. In the inference phase, the exploration term in PUCT is removed (i.e., $c_{puct}=0$), and node selection relies primarily on value estimates from the trained value model for both intermediate nodes and candidate answer nodes, focusing the search on promising branches and improving efficiency. This design is motivated by the fact that data analysis tasks have a vast, sparse search space, where most branches are invalid, while correct solutions are concentrated in a few high-quality branches. Retaining the exploration term causes unnecessary exploration and computation. In contrast, a well-trained value model effectively identifies promising nodes, enabling more focused and efficient search—a result also supported by our experiments.

The search proceeds until a maximum number of iterations or a stopping criterion is reached, and all candidate answer nodes encountered during the search are retained. For tasks with reference answers, the final answer can be determined by majority voting or by selecting the candidate with the highest value estimate; for open-ended tasks, other task-specific evaluation metrics such as validation set accuracy can be used.

\subsection{Value Model Training}

To train the value model, we first extract both successful and failed reasoning trajectories collected from 6,845 samples in NbQA during the Trajectory Collection phase. For each node along a trajectory, we use its corresponding context as input and the Q-value obtained from multiple MCTS simulations as the supervision signal. The value model is built upon the base language model fine-tuned on NbQA, with an additional regression-based value head attached. The value head outputs a scalar value in the range $[-1, 1]$ by pooling the final hidden states, and is trained using mean squared error loss against the normalized MCTS Q-values. Once trained, the value model provides accurate value estimation to guide tree expansion during inference, leading to more efficient and effective search. Experimental settings are provided in the supplementary materials.

\section{Experiments}

We first present the performance improvements brought by supervised fine-tuning (SFT) of several models on the InfiAgent-DABench benchmark. We then highlight the further gains achieved by value-guided search in \method, comparing it against other popular inference strategies and agent frameworks. Next, we evaluate the generalization ability of our approach on DSBench and AIME, demonstrating both the transferability of the value model across diverse data analysis task formats and the enhancement in tool-use capability after fine-tuning. For more detailed descriptions of the experimental setup and the effects of different hyperparameter values, please refer to the supplementary materials.

\subsection{Finetuning Results}

\begin{table}[t]
\centering
\begin{tabular}{c|l}
\hline
\textbf{Model} & \textbf{Accuracy by
Questions /\%}  \\
\hline
Mistral-7B-Instruct-v0.3 & \makebox[3em][l]{\textbf{+56.81}} (\phantom{0}2.33 $\to$ 59.14) \\
Llama-3.1-8B-Instruct      & \makebox[3em][l]{\textbf{+21.40}} (48.25 $\to$ 69.65) \\
Qwen2.5-7B-Instruct        & \makebox[3em][l]{\textbf{+24.12}} (43.97 $\to$ 68.09) \\
QWen2.5-14B-Instruct       & \makebox[3em][l]{\textbf{+7.39}}  (69.65 $\to$ 77.04) \\
\hline
\end{tabular}
\caption{Accuracy by
Questions on InfiAgent-DABench before and after finetuning on NbQA.}
\label{tab:finetune-results}
\end{table}

We randomly sampled 8{,}975 instances from the NbQA, ensuring that benchmark data was excluded from both the supervised fine-tuning and trajectory collection phases, to construct a multi-turn interaction fine-tuning dataset following the ReAct-style~\cite{yao2023reactsynergizingreasoningacting} format. The input prompt includes the task, constraints, format, and required data files. The model is instructed to generate a \textit{thought} and an \textit{action} for each round of iteration based on the current context. The extracted python code block in \textit{action} is executed by the Jupyter kernel of SandboxFusion \cite{bytedanceseedfoundationcodeteam2025fullstackbenchevaluatingllms}, and the resulting output is incorporated back into the history. This process continues until the model determines that sufficient information has been collected to output the formatted answer. We set the temperature to 0.2 and maximum iteration to 25, as required by \cite{hu2024infiagentdabenchevaluatingagentsdata}. 

\textbf{As shown in Table \ref{tab:finetune-results}}, After 3 epochs of LoRA finetuning \cite{hu2021loralowrankadaptationlarge}, the performance of 7B models on InfiAgent-DABench improves by more than 21\% in accuracy by questions. This demonstrates that our constructed dataset effectively enables the model to enhance data analysis knowledge and multi-turn reasoning abilities.

\subsection{InfiAgent-DABench}

\begin{table*}[!tb]
\centering
\begin{tabular}{c|>{\centering\arraybackslash}m{7.5cm}|c}
\hline
\textbf{Model} & \textbf{Framework} & \textbf{Acc. by Q. (\%) } \\
\hline
\multirow{4}{*}{GPT-4o mini}
& ReAct            & 80.08 \\
& Taskweaver       & 76.65 \\
& AutoGen          & 70.04 \\
& Data Interpreter & 67.7 \\
\hline
\multirow{4}{*}{GPT-4o} 
& ReAct            & 81.32 \\
& AutoGen          & 73.54 \\
& Taskweaver       & 85.99 \\
& Data Interpreter & 75.78 \\
\hline
\multirow{2}{*}{Qwen2.5-72B-Instruct}
& ReAct            & 75.88  \\
& AutoGen          & 70.04 \\
\hline
\multirow{6}{*}{Qwen2.5-7B-Instruct \textit{(SFT)}}
& ReAct                                 & 68.09 \\
& Majority Voting (25 Iters, 5 Runs)    & 75.10 \\
& \method~(40 Iters, \makebox[1.5em][l]{w/o} VM, \makebox[2em][l]{w/o} ExpTerm)  & 70.04 \\
& \method~(40 Iters, \makebox[1.5em][l]{w/o} VM, \makebox[2em][l]{with} ExpTerm) & 68.87 \\
& \method~(40 Iters, \makebox[1.5em][l]{with} VM, \makebox[2em][l]{with} ExpTerm) & 68.87 \\
& \method~(40 Iters, \makebox[1.5em][l]{with} VM, \makebox[2em][l]{w/o} ExpTerm)& 77.82 \\
\hline
\multirow{6}{*}{Qwen2.5-14B-Instruct \textit{(SFT)}}
& ReAct                                 & 77.04 \\
& Majority Voting (25 Iters, 5 Runs)    & 83.66 \\
& \method~(40 Iters, \makebox[1.5em][l]{w/o} VM, \makebox[2em][l]{w/o} ExpTerm)  & 79.38 \\
& \method~(40 Iters, \makebox[1.5em][l]{w/o} VM, \makebox[2em][l]{with} ExpTerm) & 75.88 \\
& \method~(40 Iters, \makebox[1.5em][l]{with} VM, \makebox[2em][l]{with} ExpTerm)& 74.71 \\
& \method~(40 Iters, \makebox[1.5em][l]{with} VM, \makebox[2em][l]{w/o} ExpTerm) & \textbf{\underline{86.38}} \\
\hline
\end{tabular}
\caption{Accuracy by questions on InfiAgent-DABench across various model settings. Models marked with \textit{(SFT)} are supervised fine-tuned on NbQA. “w/o VM” and “with VM” indicate whether a trained value model is used. “w/o ExpTerm” and “with ExpTerm” indicate whether the exploration term in the PUCT selection algorithm is disabled ($c_{puct} = 0$) or enabled ($c_{puct} = 1.25$), respectively. “Iters” denotes the maximum number of search iterations allowed per question, and “Runs” denotes the number of independently sampled candidate solutions per question.}
\label{tab:main-results}
\end{table*}

We compare \method~with several state-of-the-art agent systems and inference strategies on the InfiAgent-DABench benchmark, including Majority Voting \cite{wang2023selfconsistencyimproveschainthought}, ReAct \cite{yao2023reactsynergizingreasoningacting}, AutoGen \cite{wu2023autogenenablingnextgenllm}, Taskweaver \cite{qiao2024taskweavercodefirstagentframework}, and Data Interpreter \cite{hong2024datainterpreterllmagent}. For search and Majority Voting, we adopt the same prompt and multi-turn interaction format as used during fine-tuning. The maximum search tree depth is set to 10, with a maximum of 40 iterations. At each iteration, three sequences are sampled to expand the selected node. Each search path allows up to three code execution errors. For intermediate nodes and candidate answer nodes encountered during search, if the value model is used, value estimates are obtained from the value model and backpropagated; otherwise, a value of 0 is assigned and backpropagated. Finally, all candidate answer nodes are collected, and the final answer is determined by taking the mode among them. For Majority Voting, the temperature is set to 0.7. For each question, a maximum of 25 iterations is allowed, and five candidate solutions are sampled. For \method, the temperature is also set to 0.7. For agent frameworks using GPT-4o, GPT-4o mini, and Qwen2.5-72B-Instruct, results are reported by \cite{you2025datawiseagentnotebookcentricllmagent}, with the maximum number of iterations set to 21. The temperature for ReAct is set to 0.2, and for all other agents, the temperature is set to 0.

\textbf{As shown in Table \ref{tab:main-results}}, \method~achieves its highest accuracy on InfiAgent-DABench when both the value model is used and the exploration term is removed. In this setting, the fine-tuned Qwen2.5-14B-Instruct reaches an ABQ of 86.38\%, surpassing the best result of Taskweaver with GPT-4o, while outperforming all other open-source and proprietary agent baselines even when these use larger or commercial models like GPT-4o. The 7B model with the same \method~configuration also shows obvious improvement.

The results further show that introducing the value model significantly boosts accuracy, and removing the exploration term leads to the best performance. When the exploration term is present, the model explores more but at the cost of lower accuracy, indicating that a strong value model alone is sufficient to efficiently guide the search. In contrast, other frameworks—even with strong base models, fail to achieve this level of accuracy.

Overall, these findings suggest that value-guided search in \method~provides a general and effective means of enhancing multi-step data analysis for large language models, allowing open-source models to achieve performance competitive with the best commercial agent systems.

\subsection{DSBench}

We use the data modeling tasks from DSBench to evaluate the generalization ability of our trained value model across diverse formats of modeling tasks. All tasks focus exclusively on data modeling, without covering other types of data analysis. The data modeling tasks in DSBench are adapted from Kaggle competitions, where models are required to train machine learning models on provided training data, make predictions on the test set, write the predictions to a submission file, and compare the submission file to the ground-truth answer files to compute the task score. This process is quite different from the question-answering format used during the fine-tuning of our models. Therefore, we use vanilla QWen2.5-7B-Instruct and QWen2.5-14B-Instruct models, but still leverage our trained value model for value estimation during inference-time search. During the search, we set the maximum search tree depth to 25. For candidate answer files collected during the search process, as long as there exists at least one file that can pass the evaluation, the task is considered solved. 

Additionally, we first prompt the model itself to remove redundant sections such as platform introductions, acknowledgements, and lists of companies and organizations from the original Kaggle competition task descriptions, retaining only the key information required for the task. Experiments show that this step significantly improves performance on data modeling tasks. With the help of search, small models can even surpass the task completion rates of previous frameworks that use GPT-4 or GPT-4o as the base model, such as Code Interpreter \cite{openai-code-interpreter} and AutoGen. This additional finding demonstrates that most of the challenge in this benchmark lies in the excessive and redundant context of the direct task descriptions, rather than in the inability of small models to solve the data modeling tasks themselves. Details of the simplification approach, experimental setup are provided in the supplementary materials.

\textbf{As shown in Figure~\ref{fig: DSBench}}, with an increasing number of iterations, the task completion rate of the vanilla model without value model assistance plateaus, whereas the vanilla model with value model-assisted search continues to improve. Finally, when the number of iterations reaches 50, QWen2.5-14B-Instruct with value model-assisted search achieves a task completion rate of 98.65\%, while QWen2.5-7B-Instruct achieves 89.19\%, both surpassing QWen2.5-14B-Instruct without value model assistance. These results demonstrate that even without SFT, our trained value model can effectively assist search in different data analysis tasks, showing transferability and generalization capability.

\begin{figure}[!tb]
\centering
\includegraphics[width=1\linewidth]{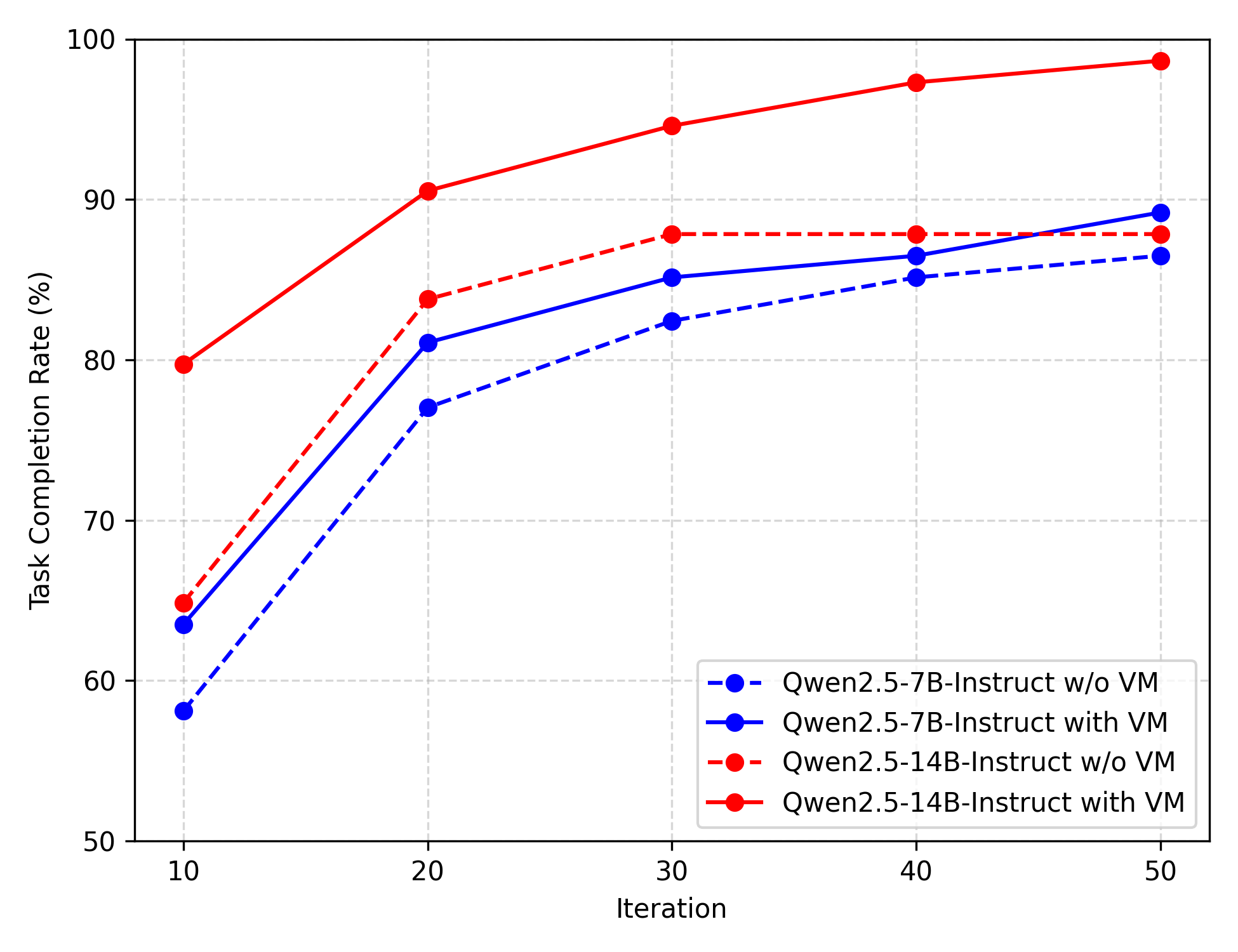}
\caption{Task completion rates of QWen2.5-7B-Instruct and QWen2.5-14B-Instruct on the DSBench data modeling task as the number of iterations increases. “w/o VM” and “with VM” denote whether a trained value model is used. For comparison, when the maximum number of iterations is limited to 40, the task completion rates of QWen2.5-7B-Instruct and QWen2.5-14B-Instruct using ReAct are 63.51\% and 66.22\%, respectively.}
\label{fig: DSBench}
\end{figure}

\subsection{AIME}

We further evaluate our models on the AIME 2025 benchmark, which consists entirely of math competition problems and lies outside the model’s training distribution. Notably, \method~remains unadapted for math-specific tasks and continues to use multi-turn tool-use prompts designed primarily for data analysis. 

\textbf{As shown in Table~\ref{tab:aime2025}}, the vanilla Qwen2.5-7B-Instruct model demonstrates extremely limited tool-use ability: the accuracy for solving with Python code execution is 0.00\%, consistent with results reported in prior work \cite{mai2025agentrlscalinglaw}. After SFT on, the \method~search achieves a voting accuracy of 13.3\% on AIME2025, which matches the result of CoT with multiple sampling before SFT. This suggests that SFT on NbQA meaningfully enhances the model’s multi-step tool-use and numerical reasoning abilities even in a fully out-of-domain math context. At the same time, the proportion of questions for which at least one correct candidate is found rises to 26.7\%, higher than CoT multi-sampling. Introducing a value model to guide the search further increases this metric to 33.3\%. This indicates that, even in completely out-of-domain settings, the value model can generalize to help the model discover more correct solutions among the candidates, increasing answer diversity, even if these correct answers are not frequent enough to dominate voting due to the SFT model limitations.

\begin{table}[!tb]
\centering
\small
\begin{tabular}{ccc}
\toprule
\textbf{Model} & \textbf{Strategy} & \textbf{Accuracy (\%)} \\
\midrule
\multirow{3}{*}{Vanilla} 
  & PoT & 0.00 \\
  & CoT & 6.67 \\
  & CoT + Multiple Sampling & 23.3 (OR) / 13.3 (Vote) \\
\midrule
\multirow{3}{*}{SFT} 
  & ReAct & 10.0 \\
  & \method~(w/o VM) & 26.7 (OR) / 13.3 (Vote) \\
  & \method~(with VM) & 33.3 (OR) / 20.0 (Vote) \\
\bottomrule
\end{tabular}
\caption{
Accuracy comparison on AIME2025 between vanilla and finetuned QWen-2.5-7B Instruct models.
"CoT" refers to using step-by-step reasoning without code execution.
"PoT" denotes solving the problem via Python code execution.
"OR" means a question is considered correct if any of the sampled answers is correct;
"VOTE" refers to majority voting. “w/o VM" and "with VM" indicate whether a trained value model is used. For CoT Multiple Sampling, 5 samples are generated for each question. For \method, the maximum search depth is set to 8, and the maximum number of iterations is 40.
}
\label{tab:aime2025}
\end{table}

\section{Related Work}

\subsection{LLM-Based Agents for Data Analysis}

Large language models have shown strong potential in automating a range of data science tasks, with encouraging progress reported in areas such as feature engineering  \cite{hollmann2023largelanguagemodelsautomated}, data visualization \cite{yang2024matplotagentmethodevaluationllmbased}, and model selection \cite{shen2023hugginggptsolvingaitasks}. Nonetheless, these efforts often target isolated stages of the data analysis workflow, overlooking the interconnected nature of data science processes and limiting their ability to provide comprehensive, end-to-end support. Several frameworks, including Taskweaver \cite{qiao2024taskweavercodefirstagentframework}, AutoGen \cite{wu2023autogenenablingnextgenllm}, and Data Interpreter \cite{hong2024datainterpreterllmagent}, have aimed to offer more integrated solutions. However, most of these frameworks rely on complex deployments and are typically based on closed-source commercial large language models and intricate task decomposition and system design. In practice, the data analysis capabilities of current language models are often limited or underestimated, particularly for complex, multi-step scenarios. To overcome these limitations, we introduce an automated pipeline and the NbQA dataset to extract and standardize high-quality, executable multi-step data analysis tasks from Jupyter notebooks. We further present \method, a value-guided search framework designed to enhance the multi-step data analysis capabilities of large language models.

\subsection{Value-Guided Tree Search}

Value-guided tree search, especially in the form of Monte Carlo Tree Search (MCTS), is a well-established method for sequential decision-making and has been widely used in areas such as game playing~\cite{silver2016mastering}, program synthesis \cite{brandfonbrener2024vermctssynthesizingmultistepprograms}, and more recently, complex reasoning with large language models~\cite{yao2023treethoughtsdeliberateproblem, _wiechowski_2022}. Recent advances like AlphaMath \cite{chen2024alphamathzeroprocesssupervision} integrate value models into the search process to improve reasoning and solution quality. These approaches directly inspired us to apply MCTS to notebook-style, multi-step data analysis workflows. Specifically, we use MCTS-generated search trajectories to train value models, which are then used at inference time to efficiently guide the search for candidate solutions. This integration enables models to achieve robust multi-step reasoning and tool-use performance on real-world data analysis tasks.

\section{Conclusion}

In this work, we construct both a dataset and a framework to enhance the data analysis capabilities of LLMs. We introduce NbQA, a large-scale dataset of real-world data analysis tasks with multi-step solutions, extracted and standardized from Jupyter notebooks to support both supervised fine-tuning and value-based learning. Building on this, we propose \method, a framework that formulates data analysis as inference-time value-guided search, enabling more accurate and efficient multi-step reasoning under limited interaction steps. Experimental results on InfiAgent-DABench and other benchmarks show that our approach significantly boosts the performance of open-source models, enabling them to match or exceed proprietary models such as GPT-4o. Together, NbQA and \method~offer a scalable and practical solution for empowering LLMs with advanced data analysis capabilities.

\bibliography{aaai2026}

\clearpage

\section*{Supplementary Material}

This document serves as supplementary material to the main paper. It mainly includes the construction process and detailed contents of the \textbf{NbQA} dataset and additional information on the experimental settings, all of which are not fully described in the main text. These details are provided to further support our research findings.

% ----------- Supplementary Content Starts Here -----------

\section{NbQA Dataset Construction Details}

\subsubsection{Crawl Notebooks on GitHub} 

We use the GitHub API \cite{githubrestapi} to crawl Jupyter Notebook files with the \texttt{.ipynb} extension from open-source code repositories. We then use regular expressions to identify data loading function calls (such as \texttt{pd.read\_csv}, etc.) within these notebooks, thereby extracting the list of data files required during notebook execution. We exclude notebooks that define file names using static variables within data loading functions. 

Based on the extracted data file lists, we directly search for and download these files from the working directory of each repository. For cases where the file path is a URL, we further examine the code comments and adjacent Markdown cells to extract download links for the corresponding data files. In total, we collected approximately 1.6 million notebooks and 3.2 million raw data files from about 47,000 GitHub repositories. We also retain notebooks whose data dependencies cannot be fully resolved, as these notebooks can still provide task-solution pairs for supervised fine-tuning.

\subsubsection{Coarse Filtering}

To ensure data quality and reliability, we first validated all crawled Jupyter Notebooks by inspecting their underlying JSON structures. We filtered out notebooks that were not executed sequentially by code cells, contained unexecuted cells, or had unhandled execution errors. To prevent data contamination and test set leakage, we removed all notebooks referencing common educational or competition datasets by performing keyword matching on both file names and content. During the experiment, we also conducted checks to ensure that the data in the involved benchmark tables did not appear in the supervised fine-tuning and trajectory collections phrase.

The filtered common educational or competition datasets include:

\begin{itemize}
    \item \textbf{Toy / Educational}: iris, boston, wine, wine recognition, breast cancer, wisconsin
    \item \textbf{Classic Machine Learning}: titanic, house prices, house-prices, adult, diabetes, heart disease, student performance, car evaluation, credit card fraud
    \item \textbf{Competition / Industry}: santander, santander customer, santander value, icr, icr-identifying, icr identifying
    \item \textbf{Image}: mnist, fashion\_mnist, cifar, imagenet, coco
    \item \textbf{Audio / Speech}: librispeech, voxceleb, urban sound, common voice
    \item \textbf{Music}: million song, fma, gtzan
    \item \textbf{Time Series / Sensor}: air quality, bike sharing, electricityload, webscope, numenta
    \item \textbf{NLP / Text}: 20newsgroups, imdb, amazon reviews, quora question pairs, ag news, sst, sms spam, trec, yahoo answers, openml, squad, sentiment140
\end{itemize}

Additionally, we excluded notebooks with overly simple tasks, such as those processing data files with fewer than 20 rows, or those with fewer than 40 lines of code in total.

To assess the quality of the notebooks, the remaining notebooks were converted to Markdown format using the nbconvert \cite{nbconvert} tool. For each successfully converted notebook, we invoked GPT-4o mini to perform an automated evaluation of both quality and suitability, assigning a score from 1 to 5 (higher scores indicate higher quality). The evaluation criteria included: whether the notebook contains well-structured, syntactically correct, and non-template code; whether the Markdown is clear and consistent with the code and its outputs; whether the dataset used is real, non-trivial, and presents meaningful challenges; whether the analysis is appropriate for the data characteristics; and whether standard Python libraries are used appropriately without relying on external APIs or high computational resources. Only notebooks with a score of at least 3 were retained. The temperature used when generating questions is set to the default value, while the temperature used when generating solutions is 0.6, and all other temperatures are set to 0. For each input, if it fails to produce an output after three attempts, the input is discarded.

\begin{tcolorbox}[colback=gray!5!white,colframe=gray!60!black,breakable,title={Prompt Used for Notebook Quality Scoring}]

You are an experienced data scientist. Please carefully review the following Jupyter notebook (converted into Markdown format using nbconvert) and determine whether it qualifies as a high-quality data analysis notebook suitable for training large language models to improve their data analysis capabilities.

Your judgment should be based on the following criteria:

- Are the Python codes syntactically correct, well-structured, and free from boilerplate or placeholders?

- Are the markdown texts clearly written and aligned with the code and results?

- Does it use real-world datasets rather than toy datasets like `iris`, `titanic`, `wine`, or synthetic/intermediate/sample data?

- Is the dataset reasonably complex and large enough to support meaningful analysis? For example: If the dataset used contains **fewer than 20 rows**, the sample size is too small to support any meaningful data analysis and should be considered low quality.

- Does it make reasonable use of common Python libraries (such as pandas, numpy, matplotlib, seaborn, sklearn, scipy, statsmodels, xgboost, keras, nltk, plotly, lightgbm, etc.) to perform data analysis?

- Notebooks must avoid relying on external APIs, pre-trained local or online models, or computation-heavy operations that require substantial CPU/GPU resources.

- Does it present a data analysis that matches the dataset and task at hand? 

- If machine learning is involved, does the notebook report key performance metrics that demonstrate practical usability, and is the training and evaluation process reproducible. i.e., can same metrics and evaluation results be obtained with the same data and configuration?

---

Please return your output in **valid, machine-readable JSON format** as shown below.

\{

    "score": an integer from 1 to 5 indicating the notebook's overall quality, with higher scores representing better quality,

    "reasoning": An explanation of why you rated this notebook as high (or low) quality and how you determined the score.

\}

---

The notebook content is as follows:

\{script\_text\}

---

Important Instructions:

All fields must be included.

Do NOT wrap your JSON output in triple backticks or Markdown code block markers. Just return valid raw JSON.
All output must be inferred from the notebook content provided above. Do not hallucinate or make assumptions beyond the content.
\end{tcolorbox}

To maximize data diversity, we retained at most one notebook per repository when the required data files were incomplete, and at most two notebooks per repository when all data files were available. Furthermore, we grouped notebooks within each repository based on the hash values of their data dependency files, and retained at most two notebooks per group.

Finally, we filtered out all notebooks involving neural networks, pretrained models, or GPU computation using keyword search, in order to focus on classical data analysis tasks.

\subsubsection{Fine-grained Processing}

To construct a thematically focused and structurally consistent dataset, we first employed GPT-4o mini to identify the machine learning models used in each notebook. For notebooks involving machine learning models, we retained only those that featured one or more of 21 commonly used classical models (such as logistic regression, support vector machines, linear regression, etc.), covering key tasks including classification, regression, clustering, and time series analysis. Approximately 54.3\% of the notebooks did not involve any machine learning models. Among all notebooks that did include machine learning models, over 81.8\% used only the aforementioned models. The full list of models, grouped by task type, is as follows:

\begin{itemize}
    \item \textbf{Classification:} Logistic Regression, Decision Tree, Random Forest, SVM, KNN, XGBoost, AdaBoost, CatBoost, LightGBM, Naive Bayes, LDA
    \item \textbf{Regression:} Linear Regression, Ridge Regression, Lasso Regression, Decision Tree, Random Forest, SVM, XGBoost, AdaBoost, CatBoost, LightGBM
    \item \textbf{Clustering:} KMeans, DBSCAN, Hierarchical Clustering, GMM
    \item \textbf{Time Series Analysis:} ARIMA, SARIMA
    \item \textbf{Dimensionality Reduction:} PCA, LDA
\end{itemize}

\begin{tcolorbox}[colback=gray!5!white,colframe=gray!60!black,breakable,title={Prompt Used for Machine Learning Model Classification}]

You are an expert code analyst and data scientist. Your task is to analyze a Jupyter Notebook that has been converted to markdown using nbconvert, and determine whether it involves the construction of any machine learning or deep learning models. Then identify all model types used in the notebook from a predefined list.

---

\#\#\# Task

Please read through the notebook content provided below and answer the following question in **JSON format**:

- Does the notebook involve constructing any **machine learning or deep learning models**?

- What are the types of models used in the notebook from the predefined list below? If multiple models are used, list all of them.

\#\#\# Predefined List of Model Types:

- Logistic Regression

- Decision Tree

- Random Forest

- SVM

- KNN

- XGBoost

- AdaBoost

- CatBoost

- LightGBM

- Naive Bayes  

- Linear Regression  

- Ridge Regression  

- Lasso Regression

- KMeans  

- DBSCAN 

- Hierarchical Clustering  

- GMM

- PCA  

- LDA

- ARIMA

- SARIMA

Use **Other** if the model type is not in the list. If no models are used, return** an empty list** for model types.

All models listed below are considered valid machine learning models, including unsupervised ones like PCA, KMeans, DBSCAN, and GMM, even if they do not use labeled data.

\#\#\# Output Format

Please return your output in **valid, machine-readable JSON format**. Use one of the following structures:

\{
    
    "thoughts": "Brief explanation of your reasoning.",
    
    "uses\_ml\_or\_dl": true or false,
    
    "model\_types": ["model\_type\_1", "model\_type\_2", ...]

\}

---

Here is the content of the notebook converted to markdown format:

\{script\_text\}

---

\#\#\# Important Instructions

- All fields must be included for each extracted task.

- Do NOT wrap your JSON output in triple backticks or markdown code block markers. Just return valid raw JSON.

- Do not invent anything or make assumptions beyond the content provided.
\end{tcolorbox}

Next, we predefined eight categories of data analysis tasks: Summary Statistics, Distribution Analysis, Correlation Analysis, Anomaly Detection, Data Preprocessing, Feature Engineering, Machine Learning, and Visualization. We then prompted GPT-4o, with examples, to extract 1 to 3 representative subtasks from each notebook, along with their corresponding answers and task types. To ensure answer verifiability, except for visualization tasks, we required that all numerical answers must be directly extractable from the code outputs in the notebook. We imposed strict constraints in the prompt: the extracted tasks must be clearly defined, context-independent, and cannot merely be a simple restatement of code execution results.

\begin{tcolorbox}[colback=gray!5!white,colframe=gray!60!black,breakable,title={Prompt Used for Question Generation}]

You are an experienced data scientist tasked with extracting high-quality data analysis subtasks and their corresponding answers from a Jupyter Notebook that has been converted to markdown using nbconvert. These subtasks will be used to train a model that only sees single natural language data analysis task and generates code to solve them.

---

\#\#\# Task

Analyze the data analysis flow of the notebook and extract only one **most representative and diverse** high-quality data analysis subtask that meet the following criteria:

1. Each task should be formulated as a **data analysis task** that can be answered by executing relevant code cells in the notebook, rather than just reading off results.

2. **Unless the task involves visualization**, the answer must be directly retrievable from the **output of code cells** in the notebook.

3. Avoid generating redundant or overly similar tasks. For example, do **not** extract multiple tasks that are minor variations of the same metric or evaluation.

4. Make each task **specific, constrained, and verifiable**. Specify the tables, columns, rows, analysis methods, tools, and expected output. A good task should provide enough detail to guide concrete analysis steps with minimal ambiguity.

5. Do **not** extract static questions that simply repeat notebook outputs, such as “What is the accuray on the test set?”, “Which model performs best?”. These are result summaries, not reasoning tasks. Focus instead on tasks that require analysis, exploration, or deeper insights.

6. Each task must be **independent and self-contained**. Do **not** assume that variables, dataframes, or preprocessing steps from previous tasks are known. All required data transformations, variable definitions, and context must be explicitly included in the task description, so it can be understood and executed in isolation. 

7. For each extracted task, return the following fields:

* `task`: The natural language formulation of the data analysis task.

* `answer`: The answer extracted directly from the code output. (Use **null** if the task involves visualization.)

* `concepts`: The analysis concepts involved (use the tags provided below; multiple tags allowed).

* `difficulty`: The difficulty of the task — "easy", "medium", or "hard".

**Reminder:** During training, the model will NOT see the original notebook content. It will only see the single natural language `task` and must generate code to solve it. Your subtasks must therefore be self-contained, clearly scoped, and useful for improving real-world data analysis capabilities.

---

\#\#\# Analysis Concept Tags (for the `concepts` field):

Use the following tags based on the type of analysis the task evaluates (you may assign multiple tags). Each tag is illustrated with an example:

- Summary Statistics

Example: Calculate the mean batting average of the players in the dataset. Exclude the one player with a missing batting average from the calculation.

- Distribution Analysis

Example: Assess whether the number of home runs hit by the players is normally distributed using the Shapiro-Wilk test at a significance level of 0.05. Exclude the player with a missing value for home runs. If the resulting p-value is less than 0.05, conclude that the distribution is not normal; otherwise, conclude that it is normal.

- Correlation Analysis

Example: Determine whether there is a correlation between the age and income columns in the dataset. Calculate Pearson’s correlation coefficient (r), which measures the strength and direction of the linear relationship between the two variables.

- Outlier Detection

Example: Identify any outliers in the temperature data for all cities listed in the file. A data point is considered an outlier if it falls more than 1.5 times the interquartile range (IQR) below the first quartile (Q1) or above the third quartile (Q3). Use only the IQR rule for outlier detection. Report the city or cities whose temperature values are identified as outliers.

- Comprehensive Data Preprocessing

Example: Perform comprehensive data preprocessing on the population column. Since there are no missing values, skip missing value handling. Convert the population values from strings to numeric data types. After the conversion, compute and report the mean and median population values.

- Feature Engineering

Example: Create a new feature called TotalCompensation by adding the BaseSalary and Bonus columns together. Then, calculate the mean TotalCompensation separately for employees who received a promotion and those who did not.

- Machine Learning

Example: Apply the linear regression algorithm from the sklearn library to predict whether an employee left the company, using the features 'Department', 'Gender', 'Age', 'YearsAtCompany', 'JobSatisfaction', and 'MonthlyIncome'. First, encode the categorical features 'Department' and 'Gender' using one-hot encoding. Then split the dataset into a training set (80\%) and a testing set (20\%) using train\_test\_split with random\_state=42. Train the linear regression model on the training data and evaluate its performance on the test data using the accuracy score.

- Visualization

Example: Draw a boxplot to show the fare distribution of male passengers.

---

\#\#\# Output Format

Please return your output in **valid, machine-readable JSON format**, as shown below:

[

  \{
    
    "task": "...",
    
    "answer": "...",  // or null if visualization
    
    "concepts": ["..."],
    
    "difficulty": "easy", "medium" or "hard"
  
  \},
  
  ...

]

---

Here are the data files that have been made locally available to support the notebook tasks, even if they were originally loaded from other paths or remote sources:

\{file\_names\}

---

Here is the content of the notebook converted to markdown format:

\{script\_text\}

---

\#\#\# Important Instructions

- All fields must be included for each extracted task.

- Do NOT wrap your JSON output in triple backticks or markdown code block markers. Just return valid raw JSON.

- Do not invent tasks that are not present in the notebook content.

- The answer must be directly retrievable from the code output, unless the task involves visualization.
\end{tcolorbox}

Considering that the initial extraction may result in ambiguously phrased tasks or inconsistent answer formats, we further require the model to supplement solution constraints and standardize the output format (e.g., specifying statistical methods, precision requirements, tolerance for randomness, etc.), and to output the answer in the form of @answer name[answer] to facilitate automatic evaluation. For visualization tasks, since the answer field is empty, only constraints are added in this step.

\begin{tcolorbox}[colback=gray!5!white,colframe=gray!60!black,breakable,title={Prompt Used for Constraints, Output Format and Label Generation}]

You are a detail-oriented data analyst. Your task is to optimize the given task by **adding constraints and output format requirements** to make the task clearer, more verifiable, and easier to assess.

---

\#\#\# Task

Please keep the **original task content unchanged**, but apply the following enhancements:

1. **Constraints**  
   
   Add detailed and precise constraints so the question has a **clear and unambiguous answer**. Ensure students do not need to make assumptions about statistical methods, parameters, or interpretation thresholds.
   
   - If the expected answer includes floating-point or scientific notation values with excessive precision, you may **reduce the precision** (e.g., from 12 decimal places to 4) to make correctness evaluation easier and avoid misleading string mismatches.
     
     - This change **must not alter the meaning** or the logic of the original answer.
     
     - You **must explicitly specify** the required number of decimal places or formatting rule in the constraints and format.

   - If the result involves **randomness** (e.g., stochastic processes, random seeds, ML model performance), and **alternative correct answers** are reasonably possible (e.g., better or slightly worse than expected accuracy). You may define a **reasonable acceptance range** (e.g., accuracy 0.8562 could be accepted as 0.86) to allow for minor variations in the answer.
    
    - You must **not simplify or reduce the difficulty** of the original problem.
    
    - The range only serves to allow **fairness in automatic evaluation**.

2. **Format**  
   
   Define an output format that allows the answer to be parsed and evaluated easily.  
   
   - Use the format `@answer name[answer]` for all answers.  
   
   - Each `answer name` should be descriptive (e.g., `@mean value[12.4]`, `@correlation coefficient[0.85]`, `@accuracy[0.87]`, `@accuracy per category[{{"A": 0.82, "B": 0.85, "C": 0.81}}]`, `@top ids[["C102", "C305", "C210"]]`).
   
   - Specify formatting rules (e.g., 2 decimal places, scientific notation, value range).

3. **Label**  
   
   Based on the original answer and your defined constraints/format, return the **final formatted answer** in a new field named `label`. 
   
   - The label must **exactly follow** the output format you've defined.

**REMEMBER** the primary evaluation method is exact string match between the `label` and the extracted answer from @answer name[answer]. If that fails, the evaluator will try converting the answer to a number and apply tolerance-based comparison; if that also fails, it will attempt to parse the value as a list and compare element-wise; if still unsuccessful, it will try to parse it as a dictionary and verify all keys match and their corresponding values meet the same comparison rules.

---

Here are some good examples for constraints, format, and label:

constraints:

Calculate the Pearson correlation coefficient (r) to assess the strength and direction of the linear relationship between height and weight. Assess the significance of the correlation using a two-tailed test with a significance level (alpha) of 0.05. Report the p-value associated with the correlation test. Consider the relationship to be linear if the p-value is less than 0.05 and the absolute value of r is greater than or equal to 0.5. Consider the relationship to be nonlinear if the p-value is less than 0.05 and the absolute value of r is less than 0.5. If the p-value is greater than or equal to 0.05, report that there is no significant correlation.

format:

@correlation coefficient[r value] @p value[p value] @relationship type[relationship type] where "r value" is a number between -1 and 1, rounded to two decimal places. where "p value" is a number between 0 and 1, rounded to four decimal places. where "relationship type" is a string that can either be "linear", "nonlinear", or "none" based on the conditions specified in the constraints.

label:

@correlation coefficient[0.48] @p value[0.0213] @relationship type[nonlinear]

---

Here are the original task-answer pair and the concepts involved:

Task: \{task\}

Answer: \{answer\}

Concepts: \{concepts\}

---

Here are the data files that have been made locally available to support the notebook tasks, even if they were originally loaded from other paths or remote sources:

\{file\_names\}

---

Here is the content of the notebook converted to markdown format:

\{script\_text\}

---

\#\#\# Output Format

Please return your output in **valid, machine-readable JSON format**, as shown below:

\{

    "thoughts": "Brief explanation of your reasoning.",
    
    "constraints": "The constraints you added to the task.",
    
    "format": "The output format you defined.",
    
    "label": "The expected final answer string, formatted according to your constraints and format.",

\}

---

\#\#\# Important Instructions

- All fields must be included for each extracted task.

- Do NOT wrap your JSON output in triple backticks or markdown code block markers. Just return valid raw JSON.

- The label field will be used for strict string-based comparison. If it fails, the evaluator will attempt to interpret the answer as a float and apply a tolerance check; if not applicable, it will try list and dictionary parsing. For lists, the comparison ensures all elements match (with tolerance if numeric); for dictionaries, all keys must exist and their values must match under the same rules. Please ensure your output is precise, structured, and suitable for automatic evaluation.

- Your "constraints" and "format" fields must NOT contain any actual answer content or directly use the value from the "label" field as a template or example. The "label" field is reserved for the final, formatted answer only. Its content must not appear in the constraints or format sections.

\end{tcolorbox}

Subsequently, we require the model to generate multi-step solutions corresponding to each extracted task. Unlike previous methods that rely on large language models to synthesize answers, our approach does not depend on the model to generate answers directly. Instead, the model is instructed to extract relevant code and outputs directly from the original notebook. We explicitly prohibit the model from generating outputs, inventing logic, or skipping steps, ensuring that each solution path (composed of multiple \texttt{thought-code-output} steps and a final \texttt{thought-label}) faithfully reflects the original analytical process. For visualization tasks, since notebooks are provided as markdown upon input, we require the model to specify the image save link in the code output field.

\begin{tcolorbox}[colback=gray!5!white,colframe=gray!60!black,breakable,title={Prompt Used for Solution Generation}]

You are a professional data scientist. Your task is to reconstruct the **step-by-step reasoning and code process** that leads to solving a real data science problem.

The `task`, `answer`, `constraints`, and `label` below were extracted from an original Jupyter notebook (converted to markdown using nbconvert). You must recover the reasoning and computation path — using the notebook's actual code blocks and outputs — that a human data scientist would follow to reach this result.

---

\#\#\# Important Guidelines

You must not modify the task, answer, constraints, or label — they are extracted from the original notebook and must remain unchanged. You are not allowed to skip steps or directly output the final result. Instead, simulate the real decision-making workflow of a data scientist who iteratively builds toward the final solution, thinking through each step based on what has already been done.

All output-producing code must use `print(...)`. Do not use `display(...)`, and do not rely on implicit output (e.g., just writing `df.head()` or `model.score(X, y)` without wrapping them in `print(...)`).  

If such implicit output exists in the original notebook, **you must rewrite it using `print(...)`**, and correspondingly update the `code\_output` to reflect what would be printed explicitly.  This ensures the reasoning path works in non-interactive environments (such as script execution or logs), where implicit outputs would otherwise be lost or misleading.

Note:

Do **not** invent new code or introduce logic that does not appear in the original notebook. Except for minimal adjustments (such as variable renaming or converting outputs to `print(...)`), your code should closely mirror the original. Because you cannot execute code or verify the actual output, avoid speculative modifications that may lead to inconsistent or inaccurate results.

---

\#\#\# Task Information

Task: \{qa["task"]\}

Answer: \{qa["answer"]\}

Constraints: \{qa["constraints"]\}

Format: \{qa["format"]\}

Label: \{qa["label"]\}

Concepts: \{str(qa["concepts"])\}

---

\#\#\# Available Files in Local Directory

\{qa["file\_names"]\}

---

\#\#\# Original Notebook Content

This is the original notebook content that has been converted to markdown using nbconvert. Extract code blocks and outputs from this to construct your solution:

\{script\_text\}

---

\#\#\# Required Output Format

Return a **list of dictionaries**, where each dictionary represents one step in the solution path:

- `thought`: A reasoning step that explains **why this specific action is taken now**, based on the current code and previous outputs. Think like an autonomous agent making incremental progress. Avoid generic phrases like “Now we import packages.” Instead, explain the purpose of each step in the context of advancing the task.

- `python\_code`: The actual code block (enclosed in ```python ... ```), as extracted and adapted from the notebook. You may apply light edits such as updating file paths or rewriting implicit output into explicit `print(...)` form.

- `code\_output`: The corresponding output produced by the `python\_code`. If the code has been changed from implicit to explicit print form, the output should be updated to match what `print(...)` would show.

- (Final step only) `label`: The formatted label derived from the final result, using the required constraints and format.

Here is an example of the expected output format:

[

  \{
    
    "thought": "Since the task involves a data-driven problem, the first step is to examine the dataset. By loading and printing the first few rows, I can understand what features are available and how they might relate to the target.",
    
    "python\_code": "```python
    
    df = pd.read\_csv('local\_file.csv')
    
    print(df.head())
    
    ```",
    
    "code\_output": "   col1  col2
    
    0     1     A
    
    1     2     B
    
    ..."

  \},
  
  ...
  
  \{
    
    "thought": "The previous computation yielded the final result, an accuracy of `0.934`. I will now format it to match the expected output format.",
    
    "label": "@test accuracy[0.934]"
  
  \}

]

---

\#\#\# Final Reminder

Only return the list of dictionaries in valid JSON format. Do **not** wrap the output in triple backticks or markdown.

\end{tcolorbox}

Finally, we employ GPT-4o mini to conduct a secondary review and categorization of the extracted results. First, we verify whether each answer matches the task, ensuring correctness and completeness, and check for excessively long or ambiguous responses. We then assess whether the task still exhibits significant result randomness, whether the constraints and output format are appropriate, and whether the answer label has been correctly converted and is suitable for automatic evaluation. Any task failing these checks—due to mismatched answers, inappropriate constraints, or incorrect labels—is immediately discarded.

To further facilitate automated evaluation, we also filter out any tasks whose answer label exceeds 150 characters or contains more than five \texttt{@answer\_name} items.

Lastly, we aggregate all tasks determined to involve significant result randomness, as well as those whose dependent data files could not be fully crawled.

Given that all subsequent experiments are conducted on text-based large language models and benchmarks that do not involve visualization, we excluded 12,976 visualization tasks from the main NbQA dataset, even though constraints and solutions were also generated for them. These visualization tasks are separated and will be made available as an additional open-source resource. All subsequent statistics and experiments reported in this paper exclude this portion of the data. After all filtering steps, the final NbQA dataset comprises 38,635 task–solution pairs, of which 6,845 are associated with complete, fully accessible data files and exhibit low randomness. Considering that, despite the addition of constraints and output format requirements, the correct answers for some machine learning–related tasks in the existing dataset may still exhibit a certain degree of randomness, we conducted an additional experiment in which all machine learning–related tasks were removed from the task set used during trajectory collection. We did not observe any significant impact of this removal on the training results of the value model or on the final search accuracy.

\begin{tcolorbox}[colback=gray!5!white,colframe=gray!60!black,breakable,title={Prompt Used for Answer Check}]

You are a detail-oriented data analyst. Your task is to evaluate a specific task and answer extracted from a Jupyter Notebook that has been converted to markdown format using nbconvert.

---

\#\#\# Task

Please carefully evaluate the task-answer pair in the context of the provided data files and notebook content. Then provide a detailed evaluation of the answer based on the following criteria:

1. **Answer Alignment:** Does the answer explicitly and directly respond to the task, rather than merely pointing out where to find it or how to derive it?

2. **Answer Verifiable:** Can the answer be verified, derived, or directly found based on the notebook code cells' output?

3. **Answer Correctness:** Is the answer free from any errors or inaccuracies?

4. **Answer Completeness:** Does the answer provide a complete and accurate response to the task rather than an incomplete or partial answer?

5. **Answer Excessive or Vague:** Is the answer excessively long, overly generic, or vague such that it becomes difficult to assess?

---

Here are the original task-answer pair and the concepts involved:

Task: \{task\}

Answer: \{answer\}

Concepts: \{concepts\}

---

Here are the data files that have been made locally available to support the notebook tasks, even if they were originally loaded from other paths or remote sources:

\{file\_names\}

---

Here is the content of the notebook converted to markdown format:

\{script\_text\}

---

\#\#\# Output Format

Please return your output in **valid, machine-readable JSON format**, as shown below:

\{

    "thoughts": "Brief explanation of your reasoning.",
    
    "answer\_alignment": "yes" or "no",
    
    "answer\_verifiable": "yes" or "no",
    
    "answer\_correctness": "yes" or "no",
    
    "answer\_completeness": "yes" or "no",
    
    "answer\_excessive\_or\_vague": "yes" or "no"
    
\}

---

\#\#\# Important Instructions

- All fields must be included for each extracted task.

- Do NOT wrap your JSON output in triple backticks or markdown code block markers. Just return valid raw JSON.

\end{tcolorbox}

\begin{tcolorbox}[colback=gray!5!white,colframe=gray!60!black,breakable,title={Prompt Used for Constraints, Output Format and Label Check}]

You are a detail-oriented data analyst. Your task is to **verify the quality of the constraints, output format, and label** that were generated for a specific task-answer pair. The task-answer pair is extracted from a Jupyter Notebook that has been converted to markdown format using nbconvert. You must assess whether the constraints, format, and label are valid, appropriate, and strictly faithful to the original task-answer pair.

---

\#\#\# Task

Please carefully assess the following three aspects:

1. **Randomness or Evaluation Difficulty**
   
   - Despite the added constraints, does the task outcome still suffer from high randomness or variability that would significantly hinder correctness evaluation, beyond minor acceptable fluctuations?
   
   - Example: A K-Means clustering task without a fixed random\_state may yield inconsistent cluster centers across runs.

2.  **Constraints and Format Appropriateness**
   
   - Are the constraints a clear, concrete specialization of the original task? Do they remove ambiguity and clarify expected value types, decimal precision, and structural requirements (e.g., list/dict format)?  
   
   - Does the format follow the @answer name[answer] convention and completely capture all required answer elements in the task and constraints?

3. **Label Soundness and Evaluability**  
   
   - Is the `label` a faithful transformation from the task and constraints?  
   
   - Does it preserve the semantic content, with only changes in formatting or precision?  
   
   - Is the label **suitable for automated evaluation**?

**REMEMBER** the primary evaluation method is exact string match between the `label` and the extracted answer from @answer name[answer]. If that fails, the evaluator will try converting the answer to a number and apply tolerance-based comparison; if that also fails, it will attempt to parse the value as a list and compare element-wise; if still unsuccessful, it will try to parse it as a dictionary and verify all keys match and their corresponding values meet the same comparison rules.

---

Here are the original task-answer pair, constraints, output format, label, and concepts involved:

Task: \{qa["task"]\}

Answer: \{qa["answer"]\}

Constraints: \{qa["constraints"]\}

Format: \{qa["format"]\}

Label: \{qa["label"]\}

Concepts: \{str(qa["concepts"])\} 

---

Here are the data files that have been made locally available to support the notebook tasks, even if they were originally loaded from other paths or remote sources:

\{qa["file\_names"]\}

---

Here is the content of the notebook converted to markdown format:

\{script\_text\}

---

\#\#\# Output Format

Please return your output in **valid, machine-readable JSON format**, as shown below:

\{
  
  "thoughts": "Brief explanation of your reasoning.",
  
  "has\_randomness\_or\_evaluation\_difficulty": "yes" or "no",
  
  "constraints\_and\_format\_appropriate": "yes" or "no",
  
  "label\_reasonable\_and\_evaluable": "yes" or "no"

\}

---

\#\#\# Important Instructions

- All fields must be included for each extracted task.

- Do NOT wrap your JSON output in triple backticks or markdown code block markers. Just return valid raw JSON.

\end{tcolorbox}

\section{Overview of the NbQA dataset}

We separately report statistics for the subset of NbQA 6,845 tasks that have complete data dependency files and exhibit low randomness—making them suitable for supervised fine-tuning (SFT) and value model training—and 31,790 tasks that either lack complete data dependency files or exhibit a certain degree of randomness.

For tasks that either lack complete data dependency files or exhibit a certain degree of randomness, the majority of processed files are in the \texttt{.csv} format, accounting for 45,679 files and representing the vast majority of the dataset. This is followed by \texttt{.xlsx} (3,298 files) and \texttt{.txt} (2,454 files), while other types such as \texttt{.parquet}, \texttt{.tsv}, and \texttt{.json} are all below one thousand. For tasks that have complete data dependency files and exhibit low randomness, the number of \texttt{.csv} files is 8,961, followed by \texttt{.txt} (401 files) and files with no extension (180 files). Other types, such as \texttt{.tsv}, \texttt{.json}, and \texttt{.data}, each have fewer than one hundred samples.

\begin{figure}[!htbp]
\centering
\includegraphics[width=1\linewidth]{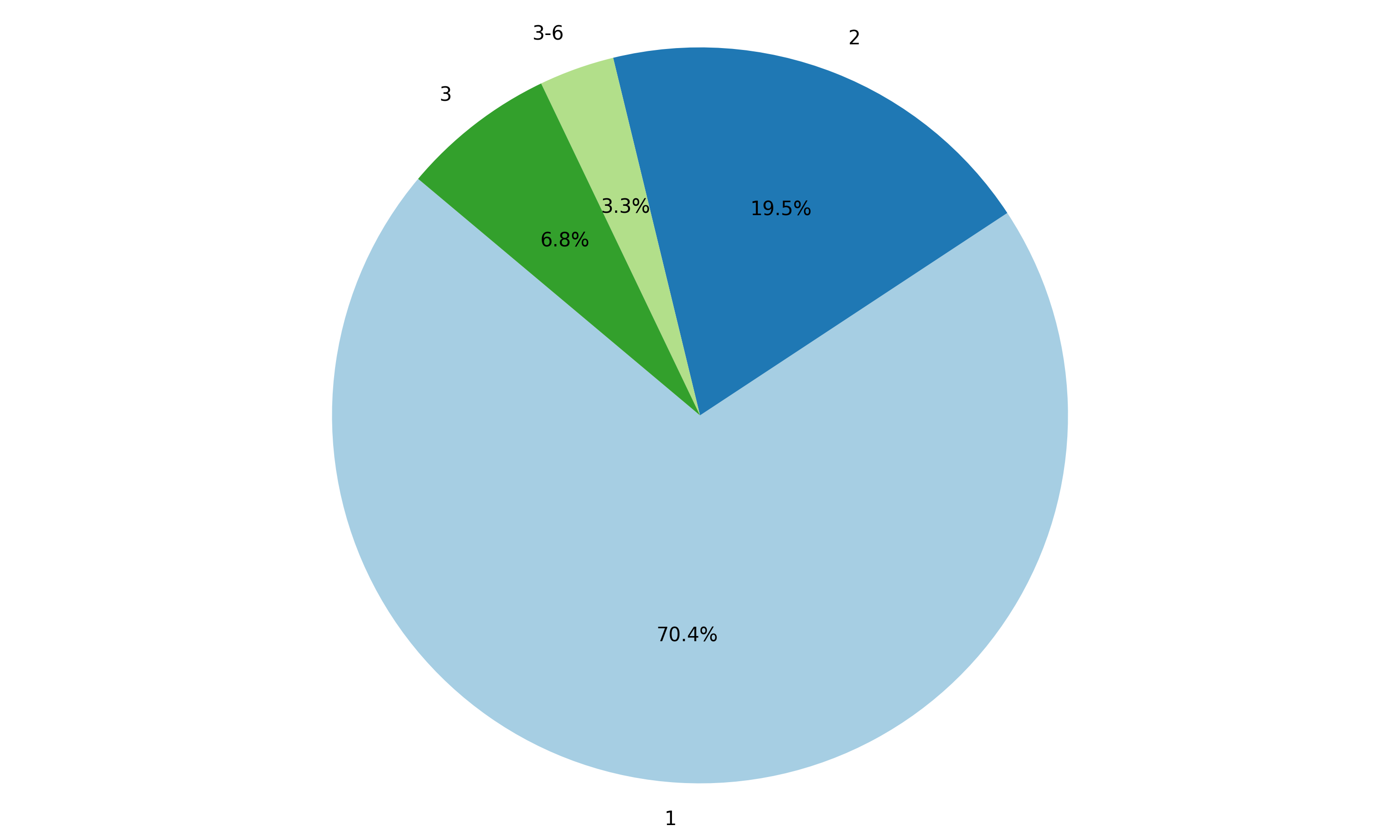}
\caption{The distribution of the number of data files processed by tasks for tasks that have complete data dependency files and exhibit low randomness}
\end{figure}

\begin{figure}[!htbp]
\centering
\includegraphics[width=1\linewidth]{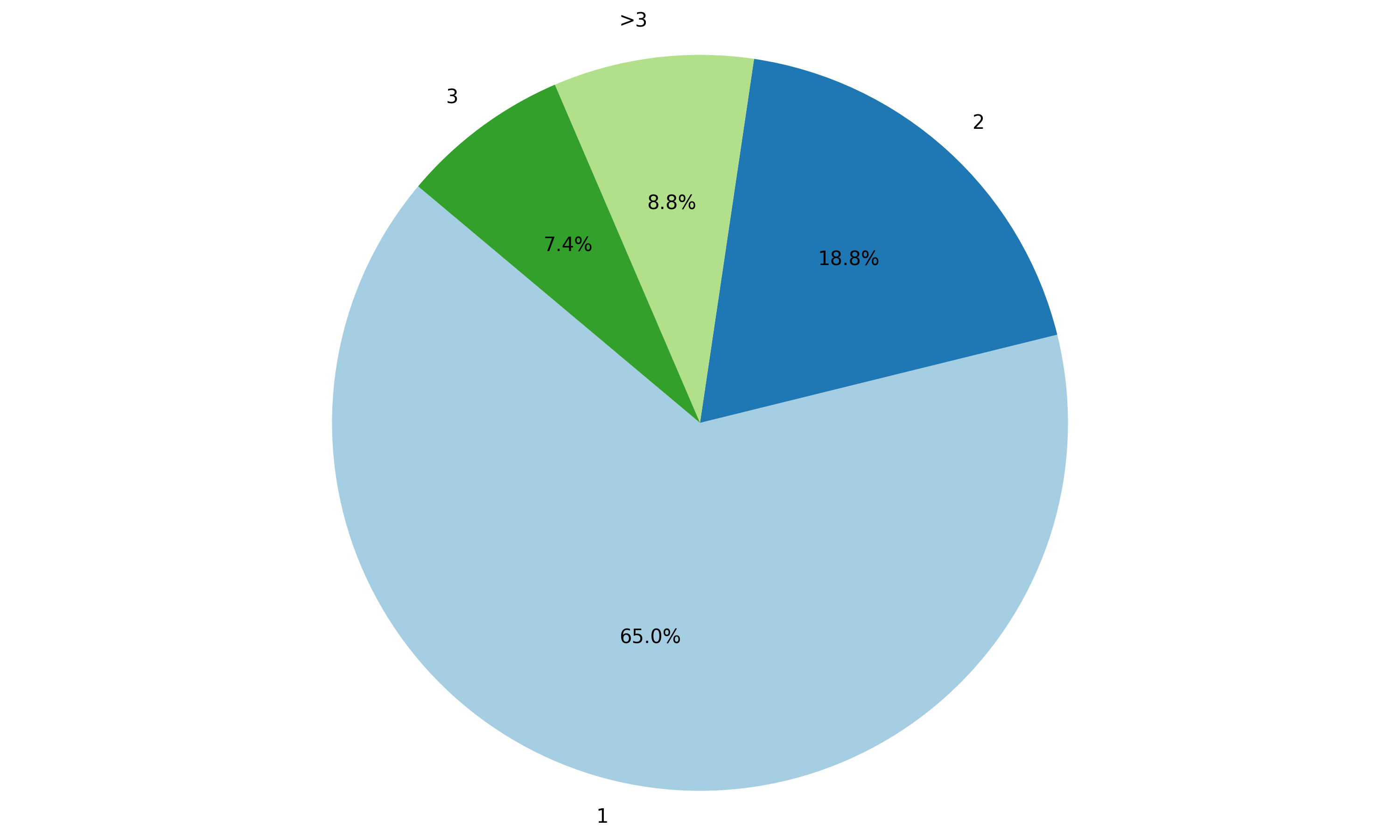}
\caption{The distribution of the number of data files processed by tasks for tasks that either lack complete data dependency files or exhibit a certain degree of randomness}
\end{figure}

\begin{figure}[!htbp]
\centering
\includegraphics[width=1\linewidth]{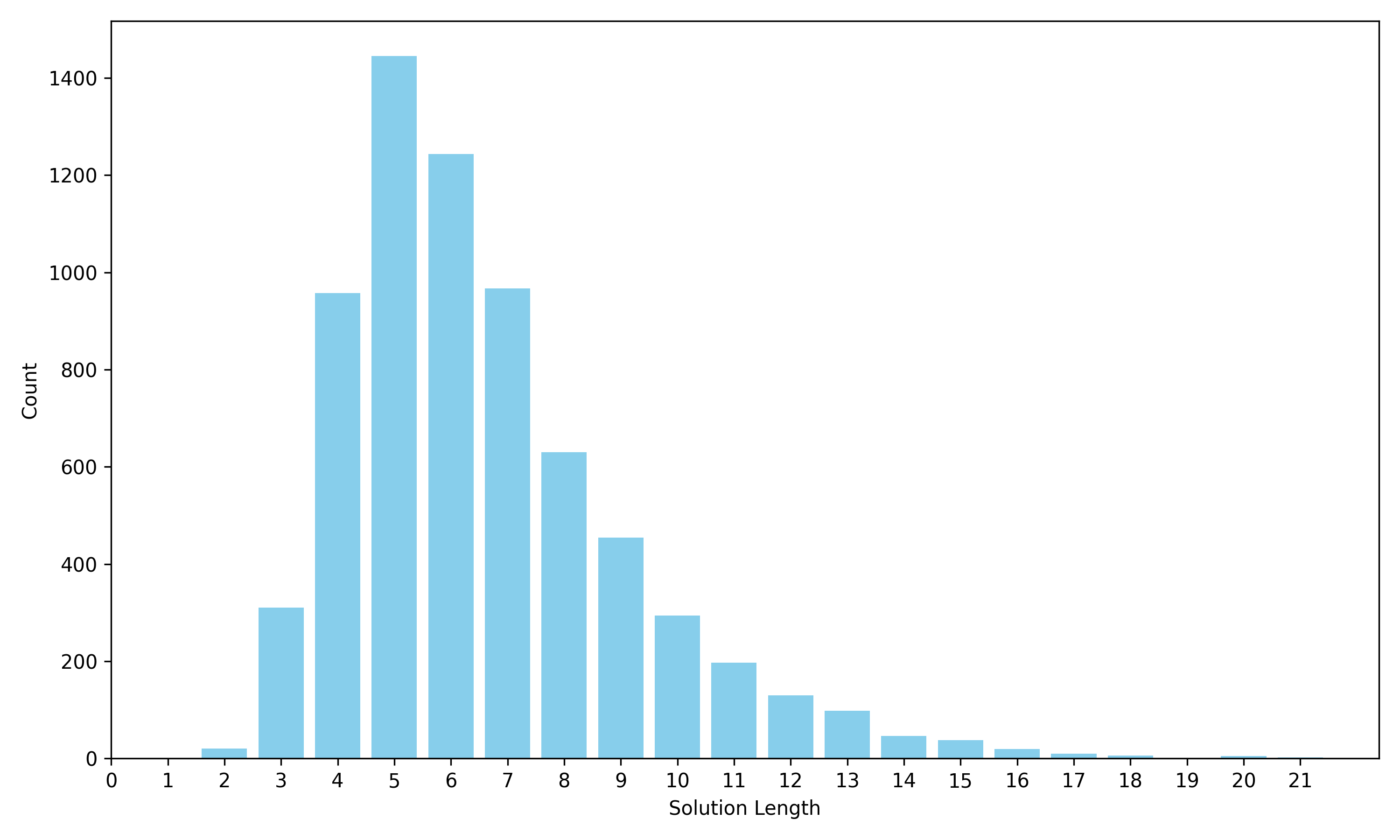}
\caption{The distribution of solution length for tasks that have complete data dependency files and exhibit low randomness}
\end{figure}

\begin{figure}[!htbp]
\centering
\includegraphics[width=1\linewidth]{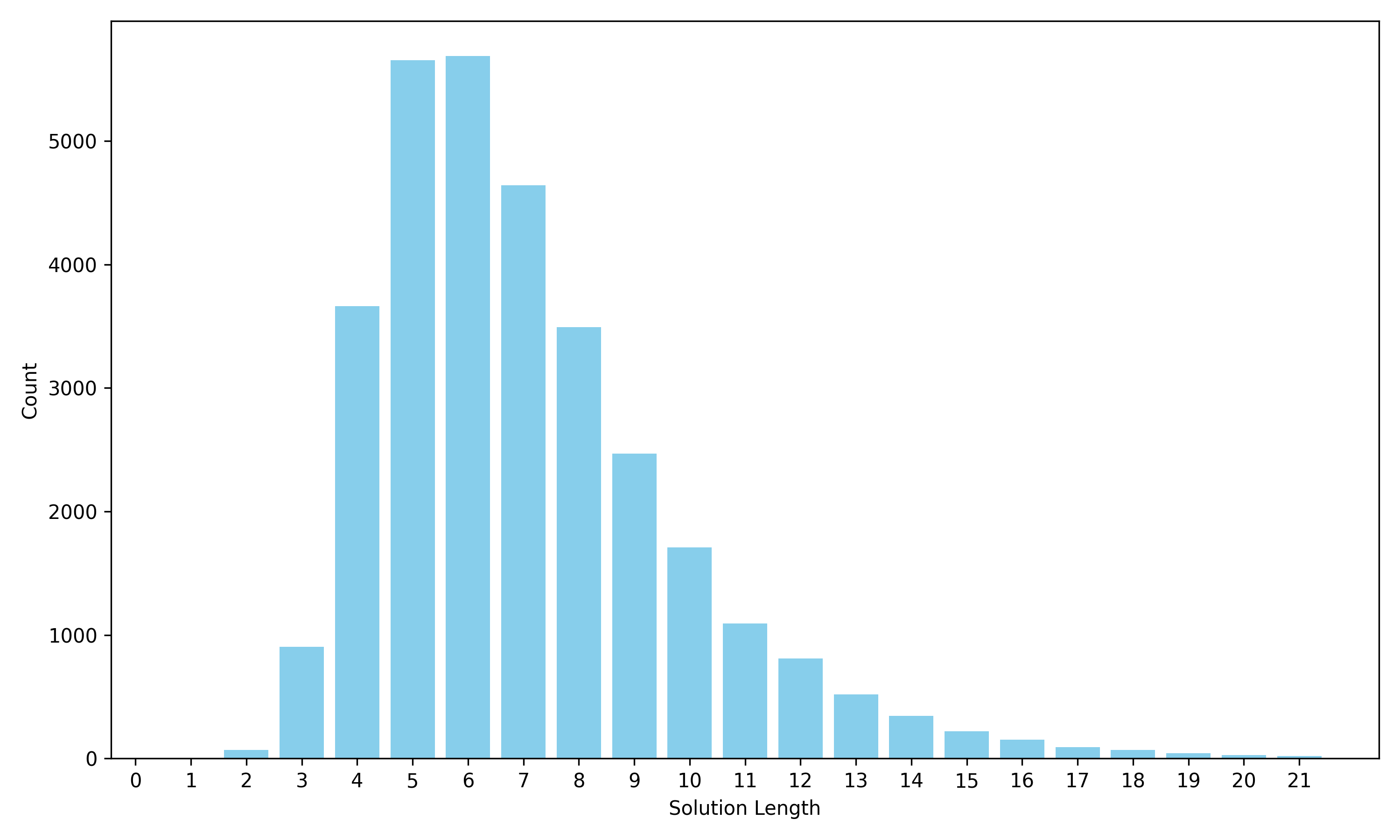}
\caption{The distribution of solution length for tasks that either lack complete data dependency files or exhibit a certain degree of randomness}
\end{figure}

\begin{figure}[!htbp]
\centering
\includegraphics[width=1\linewidth]{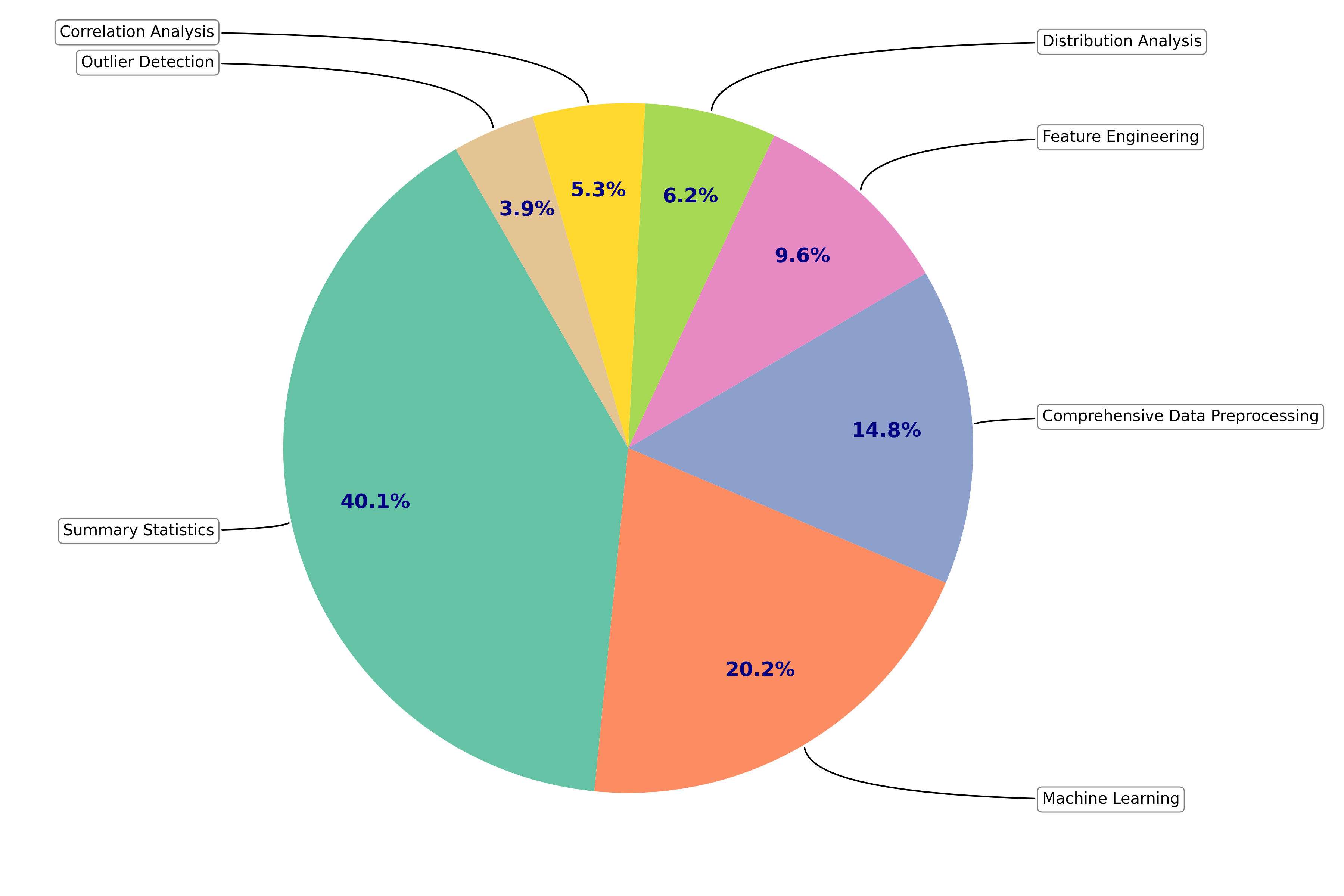}
\caption{The distribution of concepts involved in tasks that have complete data dependency files and exhibit low randomness}
\end{figure}

\begin{figure}[!htbp]
\centering
\includegraphics[width=1\linewidth]{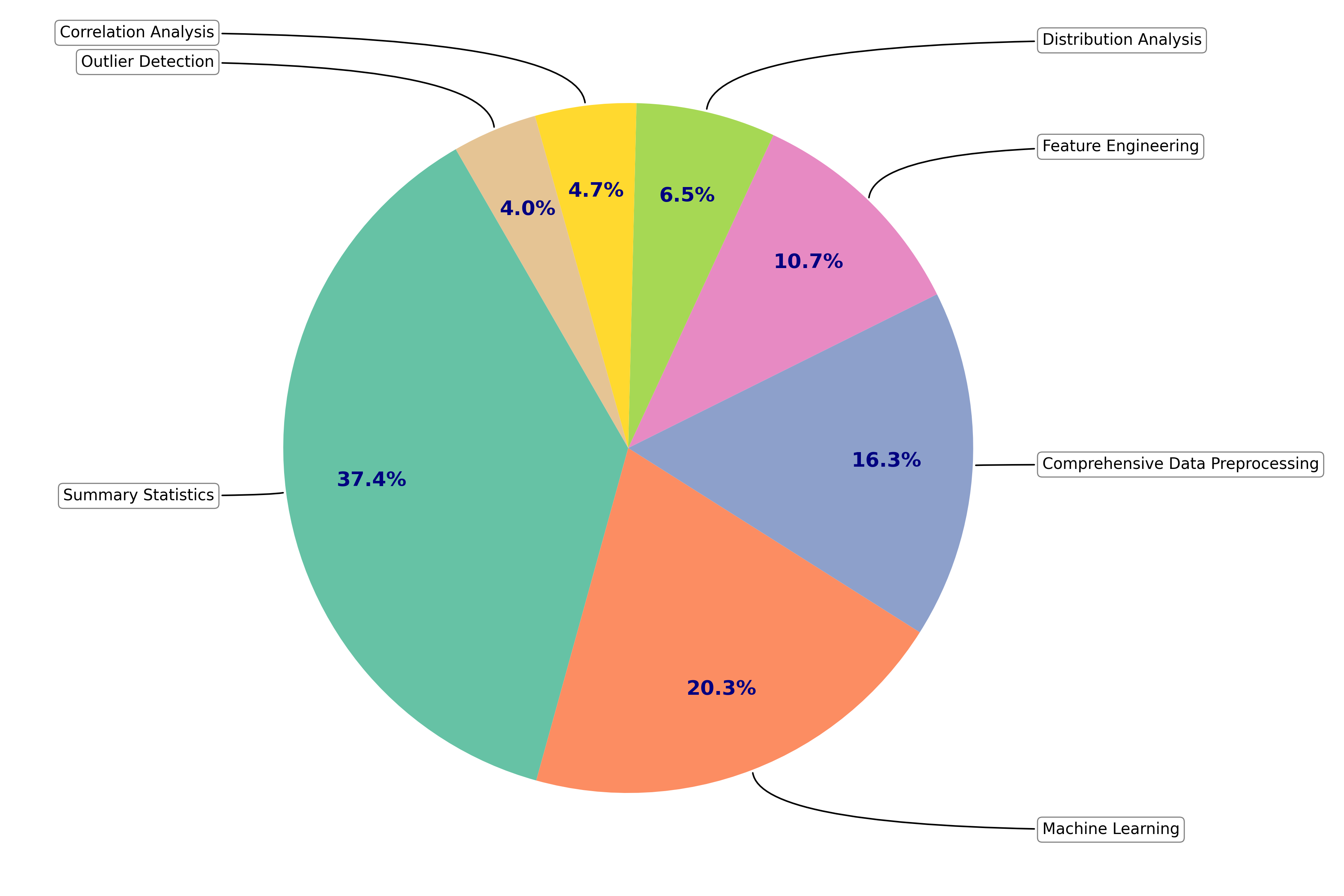}
\caption{The distribution of concepts involved in tasks that either lack complete data dependency files or exhibit a certain degree of randomness}
\end{figure}

In machine learning tasks, the most frequently used methods are Linear Regression, Random Forest, KNN, Decision Tree, SVM, KMeans, PCA, and XGBoost. These methods exhibit higher usage frequencies compared to others, while the remaining algorithms, though relatively less prevalent, still appear more than one hundred times each.

\section{Experiments Details}

\subsection{Finetuning}

We conduct supervised fine-tuning (SFT) using LLaMA Factory \cite{zheng2024llamafactoryunifiedefficientfinetuning} . Specifically, we select 8,975 samples from our dataset to construct a multi-turn finetuning dataset, where each input prompt includes the task description, constraints, required format, and necessary data files. LoRA is employed with the following settings: lora rank set to 8, lora target set to all, a cosine learning rate scheduler, a warmup ratio of 0.1, learning rate of 1.0e-4, and batch size of 16 for 3 epochs. After 3 epochs of training, we select the checkpoint at 1,500 steps for subsequent experiments.

\begin{tcolorbox}[colback=gray!5!white,colframe=gray!60!black,breakable,title={Input Prompt Used}]

Please complete the following data analysis task as best as you can. On each turn, you can use the python code sandbox to execute the python code snippets you generate and get the output.
    
\#\#\# Format to follow

On each turn, you will generate a Thought and an Action based on the current context. Here is the format to follow on each turn: 

Thought: your reasoning about what to do next based on the current context.

Action: the code snippet you want to execute in the python code sandbox. The code should be wrapped with triple backticks like ```python ... your code ... ```.

Then you will get the Observation based on the execution results from user and generate a new Thought and Action based on the new context. Once you have enough information to answer the question, finish with:

Thought: ... your final thought about the question... 

Formatted answer: the formatted answer to the original input question, should follow the constraints and format provided in the input question. 

\#\#\# Important Notes

- Do **NOT** include Observation or execution results. NEVER complete or predict the Observation part in your response. Cease generating further after generating the Action part on each turn.

- If you have any files outputted, write them to "./"

- For all outputs in code, THE print() function MUST be called. For example, when you need to call df.head(), please use print(df.head()).

Let's begin the task.

\#\#\# Task

Question: \{raw\_question\}  

Constraints: \{constraints\}  

Format: \{output\_format\}  

Available Local Files: \{file\_name\}
\end{tcolorbox}

Subsequently, we attempted to further train both the 7B and 14B SFT models with GRPO-based \cite{shao2024deepseekmathpushinglimitsmathematical} reinforcement learning, setting the input and output context lengths to 4096 and 8192, respectively. Although the reward exhibited a slight increase as the training steps progressed, the actual performance of the RL-finetuned models was inferior to that of the SFT models. Therefore, we did not adopt the RL-finetuned models in our subsequent experiments. We hypothesize that this performance degradation is mainly due to the inherent difficulty of reinforcement learning for multi-turn tool-use tasks: the model tends to repeatedly solve only simple cases, and the reward, determined by rule-based matching, is too sparse due to the varying difficulty of questions. Moreover, relying solely on rule-based checks to determine correctness may not provide sufficiently accurate feedback, especially for more complex tasks. For future work, a potential solution is to employ a model-based reward function, where a learned judge model determines whether the output is correct, thereby providing denser and more accurate reward signals for RL training.

\subsection{Trajectory Collection}

\paragraph{Collection Settings}

We collect trajectories on 6,845 tasks from the NbQA dataset, selected from 8,975 supervised fine-tuning samples. Each task has complete data dependencies and is designed for interactive execution. For trajectory collection, we use a large language model that has been previously fine-tuned on these data.

For each task, we collect a single Monte Carlo Tree (MCTS) to maximize trajectory diversity, and all trees are processed in parallel batches for efficiency. In each MCTS, the maximum number of search iterations is set to 50, and the maximum tree depth is 10. Each node expansion generates 3 candidate branches per step via LLM sampling. During LLM sampling, the temperature is set to 1 and the top-p parameter is 0.95; for each LLM call, the maximum output length is limited to 8,192 tokens. The maximum allowed input length for the LLM is 100,000 tokens; if a prompt exceeds this limit, the corresponding node is immediately terminated and assigned a negative reward. Any branch that encounters 3 consecutive code execution failures is also terminated. Each code execution in the sandbox environment has a maximum time limit of 3 minutes, with up to 40 concurrent jobs.

The reward scheme assigns $+1.0$ for fully correct final answers and $-1.0$ for failure cases such as invalid output, exceeding maximum depth, or repeated errors. In each MCTS iteration, the algorithm first selects an expandable node from each incomplete tree using the PUCT algorithm. For every node, it constructs the full message history from the root and filters out prompts that exceed the length limit. The LLM then generates 3 candidate actions per node. Each candidate is parsed for code: invalid or empty code blocks result in immediate branch termination, while valid code blocks are scheduled for parallel execution in a Python sandbox. After execution, the output—including stdout, stderr, and errors—is recorded. Branches are terminated if they reach the maximum depth or encounter too many consecutive errors. When a branch generates a final answer, it is evaluated against the ground truth, and the assigned reward is backpropagated to the root; otherwise, the branch receives zero reward.

After each tree is completed, all node information is saved to a standalone JSON file. Each node record includes its unique identifier, depth in the tree, message history, terminal status, visit counts, and value sum. 

\paragraph{Collection Results}

For the QWen2.5-7B-Instruct model, 3,004 tasks contain at least one successful trajectory, and the total number of terminal paths with positive Q-values is 73,741. During data extraction, we sampled up to 4 correct and 4 incorrect paths per task. Paths interrupted by sandbox timeouts or critical execution errors were filtered out. After deduplication and strict selection, the final dataset consists of 21,388 trajectories (including 10,853 correct and 10,535 incorrect ones), with path lengths mostly concentrated around 11 steps. For the QWen2.5-14B-Instruct model, 3,057 tasks contain at least one successful trajectory, and the total number of terminal paths with positive Q-values is 83,064. After deduplication and strict selection, the final dataset consists of 20,243 trajectories (including 11,313 correct and 89,30 incorrect ones), with path lengths mostly concentrated around 11 steps.

Each sample in the value-supervised dataset is a complete multi-turn conversation formatted as \texttt{role: content} sequences , together with a scalar Q-value that reflects the overall trajectory quality.

\subsection{Value Model Training}

\paragraph{Value Model Structure}

Our value model is constructed by attaching a regression-based value head to a language model that has already been fine-tuned on the NbQA dataset. The value head is trained to score each input prompt with a scalar value in the range $[-1, 1]$, aiming to regress the normalized Q-values produced by the MCTS search process. Specifically, each input prompt is first encoded by the base model to produce the final-layer hidden states for all tokens. We then perform mean pooling over the hidden states of valid tokens to obtain a single pooled representation. This pooled representation is fed into the value head, which outputs the final scalar score.

The \texttt{ValueHead} module consists of a linear layer with weights initialized from a Gaussian distribution, followed by a Tanh activation to restrict the output range to $[-1, 1]$ and a dropout layer to prevent overfitting. During training, the value head is optimized to fit the normalized Q-values using mean squared error (MSE) loss.

\paragraph{Training Settings}

We train the value model on collected trajectory data using SFT-pretrained Qwen2.5-7B-Instruct and Qwen2.5-14B-Instruct as base models. Training is performed for 3 epochs with a batch size of 4, a learning rate of $1 \times 10^{-4}$, and a maximum context length of 8{,}000 tokens. During training, only the LoRA adapter modules and the value head are updated, while all base model parameters are frozen. The LoRA configuration covers all attention projection modules, with hyperparameters $r=8$, $\alpha=32$, and dropout rate $0.1$. We use the AdamW optimizer with $100$ warmup steps and a weight decay of $0.01$.

\subsection{InfiAgent-DABench}

\paragraph{Experiment Settings}

For both the search method and Majority Voting with QWen2.5-7B-Instruct and QWen2.5-14B-Instruct, we adopt the same prompt and multi-turn interaction format as used during fine-tuning. For Majority Voting, the temperature is set to 0.7, with a maximum of 25 iterations per question. In each iteration, five candidate solutions are sampled, and the final answer is determined by majority vote. For the search method, the maximum search tree depth is set to 10, with up to 40 iterations allowed. At each iteration, three sequences are sampled to expand the selected node. Each search path permits up to three code execution errors, and the maximum context length is set to 100{,}000. If the value model input exceeds 8{,}000 tokens, only the most recent conversation turns are used.

For agent frameworks based on GPT-4o, GPT-4o mini, and Qwen2.5-72B-Instruct, results are taken from \cite{you2025datawiseagentnotebookcentricllmagent}. The maximum number of iterations is set to 21 for these agents. Since DataWiseAgent \cite{you2025datawiseagentnotebookcentricllmagent} has not been open-sourced, we do not include it in our comparison. In these results, the temperature for ReAct is set to 0.2, and for all other agents, the temperature is set to 0. We separately evaluate TaskWeaver with GPT-4o, setting the maximum number of turns to 21. We observe that the accuracy by questions of TaskWeaver is 82.47\%, which is lower than the 85.99\% reported in \cite{you2025datawiseagentnotebookcentricllmagent}.

\paragraph{Analysis of Search Hyperparameters}

\begin{figure}[!tb]
\centering
\includegraphics[width=1\linewidth]{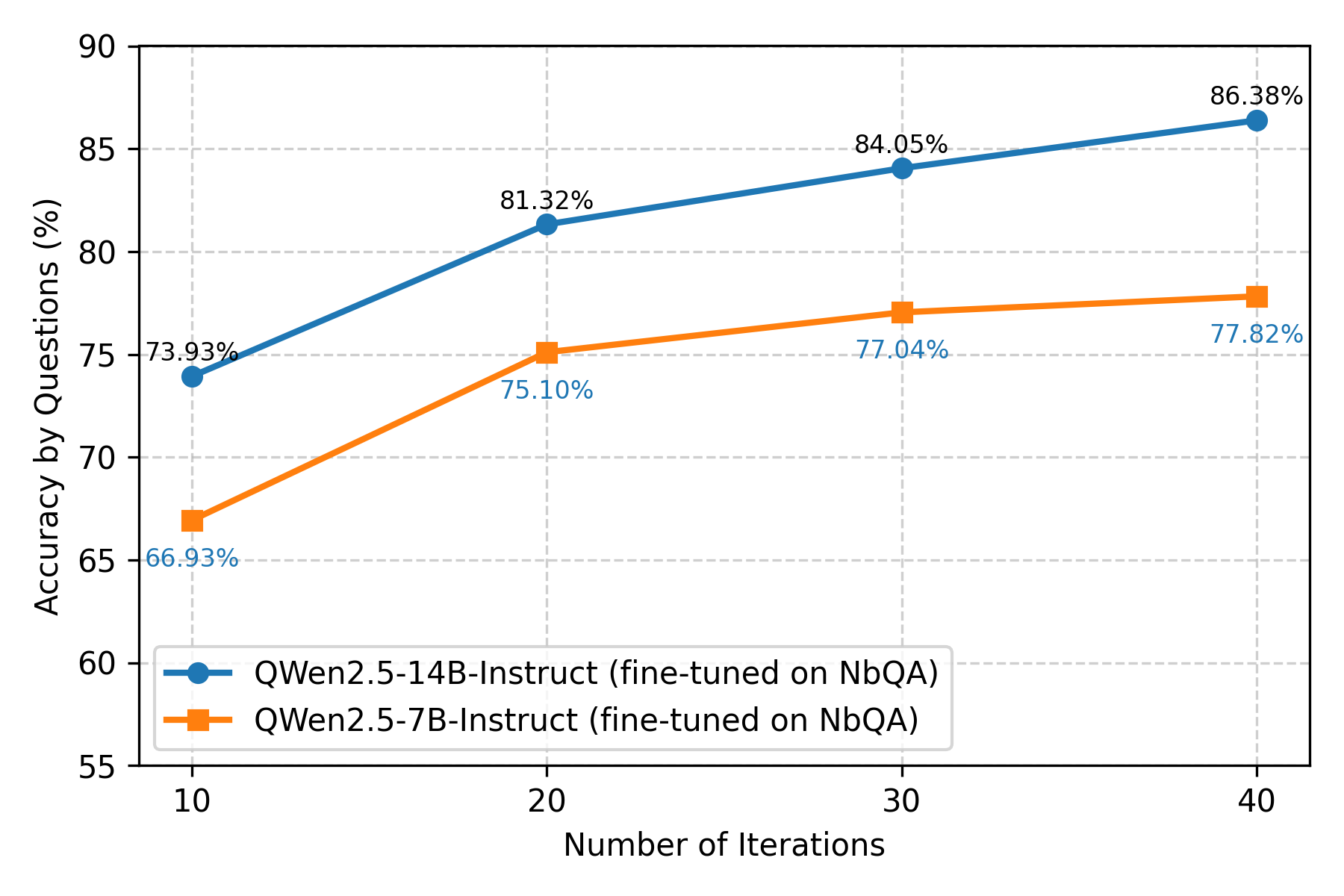}
\caption{Accuracy by questions on InfiAgent-DABench using the fine-tuned QWen2.5-7B-Instruct and Qwen2.5-14B-Instruct with the value model, MCTS exploration term removed. The sampling temperature is set to 0.7, the maximum tree depth is $10$. As the number of MCTS iterations increases, the accuracy by questions gradually improves.}
\label{fig: InfiAgent-DABench_vary_iterations}
\end{figure}

\begin{figure}[!tb]
\centering
\includegraphics[width=1\linewidth]{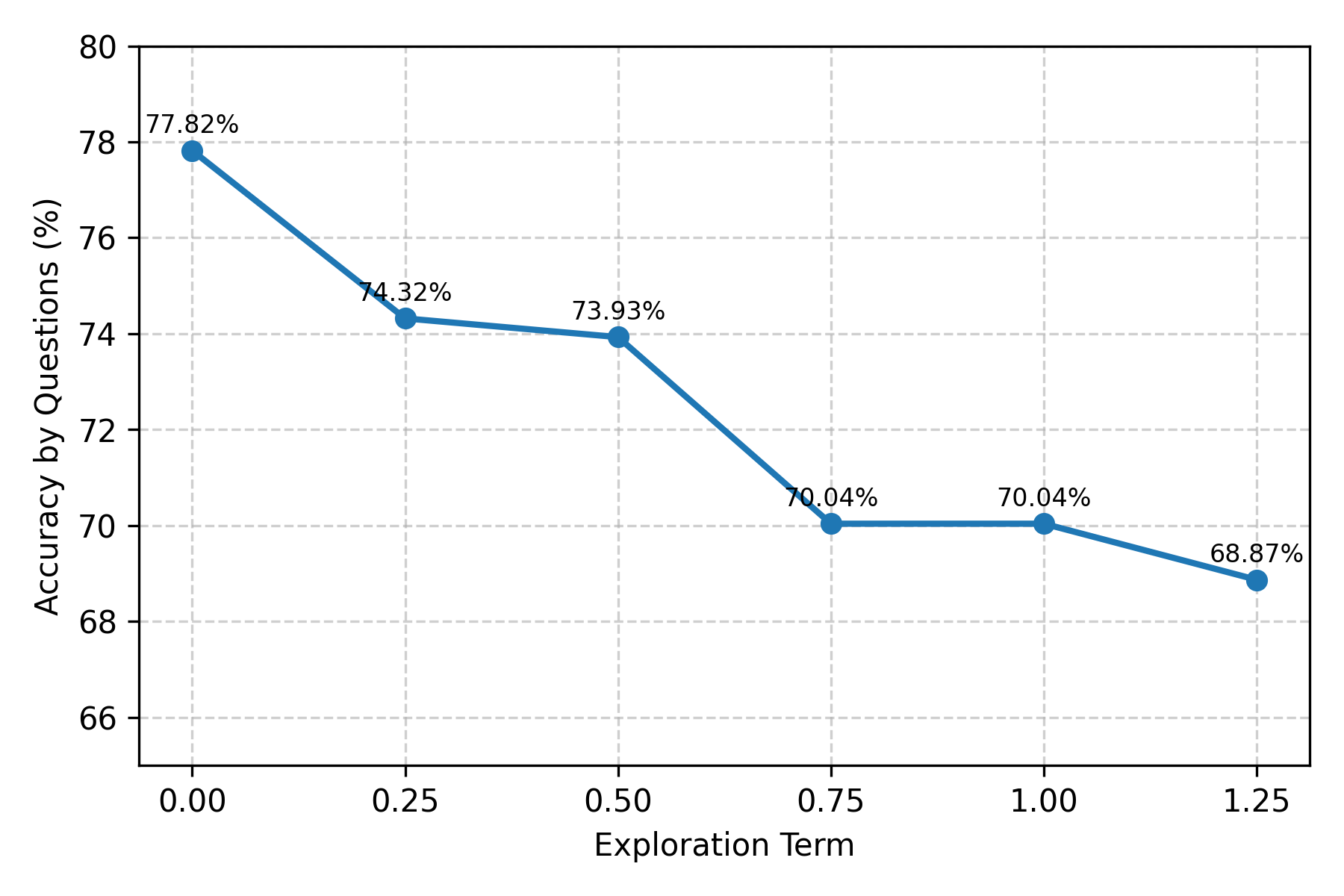}
\caption{Accuracy by questions on InfiAgent-DABench using the fine-tuned QWen2.5-7B-Instruct with the value model. The sampling temperature is set to $0.7$, the maximum tree depth is $10$. As the $c_{puct}$ increases, the accuracy by questions gradually decrease.}
\label{fig: InfiAgent-DABench_vary_exploration}
\end{figure}

\begin{figure}[!tb]
\centering
\includegraphics[width=1\linewidth]{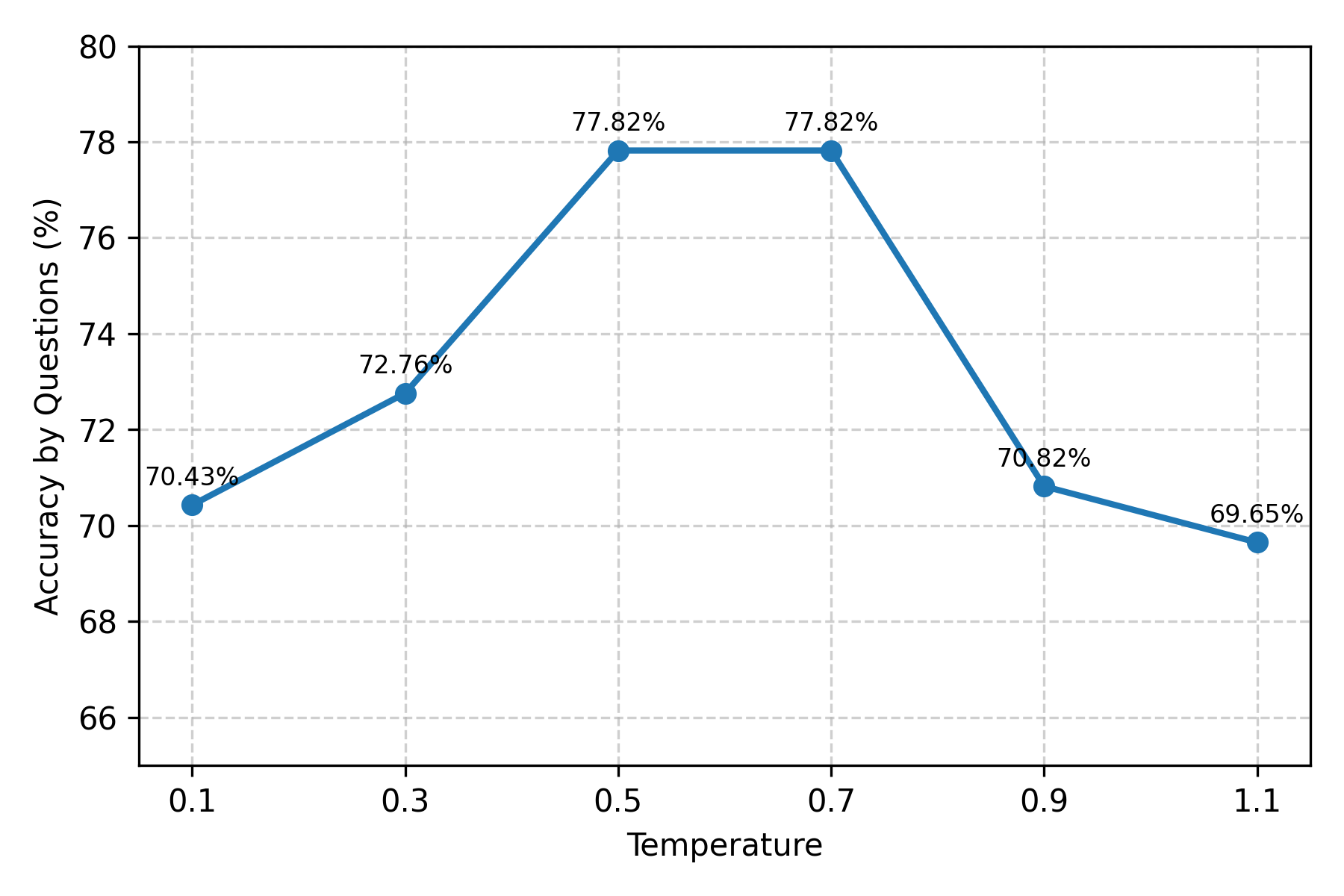}
\caption{Accuracy by questions on InfiAgent-DABench using the fine-tuned QWen2.5-7B-Instruct with the value model, MCTS exploration term removed. the maximum tree depth is $10$. As the temperature increases, the accuracy by questions first rises and then drops, with the optimal value ranging from 0.5 to 0.7.}
\label{fig: InfiAgent-DABench_vary_temperature}
\end{figure}

We also vary the values of $c_\mathrm{puct}$, the sampling temperature, and the number of iterations, and observe the corresponding changes in accuracy on InfiAgent.

\subsection{DSBench}

\begin{table*}[!tb]
\centering
\caption{The performance comparison of different models on data modeling tasks. \cite{jing2025dsbenchfardatascience}}
\begin{tabular}{llcc}
\toprule
Framework & Model & Task Success / \% & RPG \\
\midrule
\multirow{6}{*}{AutoGen}
  & Llama3-8b      & 5.41  & 1.55 \\
  & Llama3-70b     & 16.22 & 7.79 \\
  & GPT-3.5        & 8.11  & 6.02 \\
  & GPT-4          & 87.84 & 45.52 \\
  & GPT-4o         & 71.62 & 34.74 \\
  & GPT-4o mini    & 22.97 & 11.24 \\
\midrule
\multirow{4}{*}{Code Interpreter}
  & GPT-3.5        & 16.22 & 6.52 \\
  & GPT-4          & 54.05 & 26.14 \\
  & GPT-4o         & 44.59 & 19.87 \\
  & GPT-4o mini    & 39.19 & 16.90 \\
\bottomrule
\end{tabular}
\label{tab:origin_DSBench}
\end{table*}

\paragraph{Experiment Settings}

For DSBench, we also set the temperature to 0.7, with a maximum of 50 iterations per question and a maximum search tree depth of 25. Both the trained value model and the vanilla QWen2.5-7B-Instruct and QWen2.5-14B-Instruct models are used, with the same prompt, except that the format is fixed as: ``@submission\_file[submission.csv] where submission.csv is the file to be submitted for evaluation, must follow the format required in the question and constraints.'' Specifically, we first prompt the model to simplify each task description by removing redundant content from the original task and extracting the essential question and constraints as input. Experiments show that this step significantly improves performance on data modeling tasks. With the help of search, small models can even surpass the task completion rates of previous frameworks that use GPT-4 or GPT-4o as the base model, such as Code Interpreter \cite{openai-code-interpreter} and AutoGen \cite{wu2023autogenenablingnextgenllm}.

\begin{tcolorbox}[colback=gray!5!white,colframe=gray!60!black,breakable,title={Prompt Used for Simplify}]

You are an experienced data scientist. You will be given a detailed task description containing redundant information. Your job is to remove all redundant and irrelevant content, keeping only the essential information that helps a model understand and solve the task. 

Specifically, extract and concisely rewrite:

- A clear and minimal task description that explains what the problem is, what models need to do, and what the expected output is.

- Extract precise definitions of data fields in train.csv, test.csv, and sample file as well as their types or value ranges, if such information is provided in the original task description.

- A single-paragraph constraints section, containing only the necessary requirements and instructions for file usage, submission format, and output field types/values.

- Remember to remind the model to use train.csv to train a model, make predictions on test.csv, and save the results as ./submission.csv.

- Clearly specify the required content and format of the output file submission.csv in the task description or constraints, instead of referring to sample file, unless necessary.

Do not include useless background, acknowledgements, extra commentary or any other information that does not directly contribute to understanding the task.

Only output the final `question:` and `constraints:` fields.  
The output should be clear, minimal, and directly usable as task instructions for data modeling automation.

---

Please return your output in **valid, machine-readable JSON format** as shown below.

\{

    "question": "Your concise task description here.",
    
    "constraints": "Your single-paragraph constraints here."

\}

---

Here is the origin task description:

\{task\_description\}

---

Important Instructions:

All fields must be included.

Do NOT wrap your JSON output in triple backticks or Markdown code block markers. Just return valid raw JSON.

All output must be inferred from the origin task description provided above. Do not hallucinate or make assumptions beyond the content.
\end{tcolorbox}

As shown in Table \ref{tab:origin_DSBench}, When directly inputting lengthy task descriptions, almost all models exhibit low task completion rates and Relative Performance Gap (RPG) under both the Code Interpreter and AutoGen frameworks. With task simplification, when the maximum number of iterations is limited to 40, the task completion rates of QWen2.5-7B-Instruct and QWen2.5-14B-Instruct using ReAct are 63.51\% and 66.22\%, respectively. This already exceeds all other settings except for AutoGen using GPT-4o or GPT-4 as the base model. Furthermore, when equipped with search and value model assistance, the task completion rates for QWen2.5-7B-Instruct and QWen2.5-14B-Instruct reach 89.19\% and 98.65\%, respectively, which significantly outperform all cases in the table. The RPGs reach 46.04 and 51.35, indicating not only that the correct solutions are obtained via search, but also that the quality of these solutions is high. This additional finding suggests that most of the challenge in this benchmark lies in the excessive and redundant context of the direct task descriptions, rather than in the inability of small models to solve the data modeling tasks themselves.

\subsection{AIME}

For the AIME 2025 experiments, we adopt the same multi-turn tool-calling prompt as in previous settings. The temperature is set to 0.7, with three expansions per round, a maximum of 40 iterations, and a maximum tree depth of 8. The temperature of CoT Multiple Sampling is also 0.7.

\end{document}